\documentclass[10pt,journal,compsoc]{IEEEtran}

\ifCLASSOPTIONcompsoc
  \usepackage[nocompress]{cite}
\else
  \usepackage{cite}
\fi

\ifCLASSINFOpdf
  \usepackage[pdftex]{graphicx}
\else
  \usepackage[dvips]{graphicx}
\fi

\usepackage{amsmath}
\usepackage{amsfonts}

\ifCLASSOPTIONcompsoc
  \usepackage[caption=false,font=footnotesize,labelfont=sf,textfont=sf]{subfig}
\else
  \usepackage[caption=false,font=footnotesize]{subfig}
\fi

\usepackage{url}

\newtheorem{theorem}{Theorem}

\usepackage{xcolor}

\begin{document}
\title{Training Feedforward Neural Networks with\\Standard Logistic Activations is Feasible}

\author{Emanuele Sansone,
        and~Francesco~G.D.~De~Natale,~\IEEEmembership{Member,~IEEE}
}

\IEEEtitleabstractindextext{%
\begin{abstract}
Training feedforward neural networks with standard logistic activations is considered 
difficult because of the intrinsic properties of these sigmoidal functions. This work aims
at showing that these networks can be trained to achieve generalization performance 
comparable to those based on hyperbolic tangent activations. The solution consists on applying
a set of conditions in parameter initialization, which have been derived from the study of the
properties of a single neuron from an information-theoretic perspective. The proposed 
initialization is validated through an extensive experimental analysis.
\end{abstract}

}

\maketitle

\IEEEdisplaynontitleabstractindextext

%
\IEEEpeerreviewmaketitle

%
\IEEEraisesectionheading{\section{Introduction}\label{sec:introduction}}
\IEEEPARstart{D}{eep learning} has received a lot of attention
in the last decade due to the impressive performance achieved
in numerous computer vision tasks, including object detection~\cite{girshick2016region},
human action recognition~\cite{ji20133d}, image restoration~\cite{dong2016image}
and image classification~\cite{krizhevsky2012imagenet}, and in natural 
language tasks, including language modelling~\cite{jozefowicz2016exploring}, 
parsing~\cite{vinyals2015grammar}, machine translation~\cite{sutskever2014sequence} 
and speech-to-text translation~\cite{graves2014towards}.
The success of deep learning is due to the capability of transforming
input data into representations that are increasingly more abstract with depth
and in a way that resembles the brain structure of primates~\cite{kruger2013deep}.
Recent theoretical analysis provides partial confirmation on
the experimental findings obtained by deep learning models and 
show that there is an exponential advantage in terms of the complexity of
functions computed by deep architectures over shallow ones~\cite{montufar2014number,raghu2016expressive,poole2016exponential}.
\footnote{We refer to shallow models to indicate networks with
a single hidden layer, while we refer to deep models for
networks with more than one hidden layer~\cite{bengio2009learning}.}

Deep learning models are impactful in many real-world applications,
and the transfer of this technology to society has created new
emerging issues, like the need of model interpretability~\cite{doshi2017towards}.
The General Data Protection Regulation approved in 2016 by the European parliament, 
which will be effective in 2018, is a concrete example of the need to provide 
human understandable justifications for decisions taken by automated data-processing
systems~\cite{goodman2016european}. 

Research could probably be inspired by old literature in neural networks to find
better explanations about the dynamics of deep learning and provide more human 
interpretable solutions. An example of such process is found in standard logistic 
activation functions, that have been studied extensively in the past, but tend to 
be substituted by other activation functions in modern neural networks. 
To understand why this may be the case, it is important to recall
the unique properties of the logistic function and therefore analyze the reasons why 
it has been introduced in neural networks. Firstly, the standard logistic function
is biologically plausible. In fact, it is one of the best differentiable 
approximations to the leaky integrate-and-fire model, used in neuroscience
to model the spiking behaviour of biological neurons~\cite{dayan2001theoretical}.
Biological plausibility can be essential for driving deep learning research
towards the uncovering of human learning dynamics and also providing explanations
that are effectively interpreted by humans~\cite{hassabis2017neuroscience}.
Secondly, there is theoretical work showing that a family of neural networks 
provided with standard logistic activations can be equivalently converted into 
fuzzy rule-based systems~\cite{benitez1997artificial}, thus raising the possibility
to perform reasoning using fuzzy logic and potentially extract human interpretable 
explanations of predictions made by deep learning models~\cite{mitra2000neuro}.

While the standard logistic function is used extensively as activation in shallow neural
networks, it receives less attention in deep learning. A common justification
supporting this fact is that training these deep neural networks is very challenging 
due to the intrinsic properties of the standard logistic function,
like its non-zero mean output value~\cite{lecun1998efficient,glorot2010understanding} and 
its low derivative score in zero~\cite{xu2016revise}. The bounded alternative
to the standard logistic activation is the hyperbolic tangent, which allows
an easier training. Nevertheless, this function is not biological plausible and does
not have any relation with fuzzy logic. This work aims at showing that training
deep feedforward neural networks with standard logistic activations 
can be feasible through careful initialization. In particular, we derive some conditions
using information theory that are used as principled criteria for initialization.
We show through extensive experimental analysis that our conditions guarantee a better
propagation of information through the whole network and that during training no
vanishing gradients are observed, thus boosting the convergence speed of the optimizer.
The proposed initialization outperforms the other existing strategies also in terms of generalization
performance and contribute to bridge the gap between networks with standard logistic activations
and networks with hyperbolic tangents.

The rest of the paper is organized as follows. Section~\ref{sec:statistical} provides
a preliminary discussion on the statistical properties of a single neuron with standard
logistic activation function. Section~\ref{sec:information} studies the neuron from
an information-theoretic perspective and derives initialization conditions for
its parameters. 
Section~\ref{sec:vanishing}
relates the proposed conditions with the problem of vanishing gradients. Finally, 
Section~\ref{sec:experiments} validates the theory over different well-known benchmarks and
different networks.

%
{\section{Literature Review}\label{sec:literature}}
In 1986, Rumelhart et al.~\cite{rumelhart1986learning} propose the backpropagation algorithm
to train a feedforward neural network.\footnote{Even if, the author in~\cite{schmidhuber2015deep}
argues that backpropagation is dated back to the early 1960s.} 
In this seminal work, the authors use random weight initialization to break the symmetry of parameters 
and allow to perform credit assignment during training, namely knowing how to compute each weight 
contribution to the final error.
Nevertheless, the initial choice of parameters plays an important role on determining 
the generalization performance of the final trained network, as it is demonstrated by subsequent
works (see~\cite{fernandez2000comparison} for an empirical comparison among the main works of the 
period until 2000 and~\cite{zhang2004mini} for an updated summary of the related work up to 2004).
In that period, the research about initialization focused mainly on shallow architectures, motivated by the fact that 
(i) shallow neural networks are 
universal function approximators~\cite{cybenko1989approximation,hornik1989multilayer}
and (ii) deep networks
are more difficult to train than shallow counterparts~\cite{bengio1994learning}, due to 
the problem of vanishing and exploding gradients.
Authors in~\cite{lecun1998efficient} are probably among the few to propose initialization strategies
for deep learning. In particular, they use random weight initialization in combination with hand-crafted
activations for hidden neurons. We will see in the experimental section that their proposed initialization 
strategy is not particularly suited for standard logistic activations, as it is strongly affected by vanishing/
exploding gradients.

The first effective strategy to learn deep models appears in 2006~\cite{hinton2006reducing} and 
consists of splitting the learning process into two stages, called unsupervised pre-training 
and fine-tuning, respectively. 
In the first stage, an unsupervised algorithm is applied layerwise to learn increasingly 
more complex representations of the input features.\footnote{Authors in~\cite{hinton2006reducing} 
use Restricted Boltzmann machines at each layer to learn deep belief nets, while authors 
in~\cite{bengio2007greedy,poultney2007efficient} use stacked autoencoders to learn deterministic 
networks.} 
In the second stage, network parameters are updated/fine-tuned using a supervised criterion 
and gradient-based optimization. 
An explanation for this success appears later in~\cite{erhan2010does}, where the authors show
experimentally that unsupervised pre-training can be regarded as an effective initialization 
strategy for the subsequent optimization stage. 
In other words, "pre-training guides the learning towards basins of attraction of minima that 
support better generalization from the training data set"~\cite{erhan2010does}.

Unsupervised pre-training is extremely expensive from a computational perspective and
alternatives are proposed to overcome the problem of vanishing/exploding gradients. In particular, 
there are solutions, which modify the structure of neural networks with skip connections 
between hidden layers~\cite{he2016deep,srivastava2015highway,huang2016densely} or with
new normalizing layers~\cite{ioffe2015batch,ba2016layer,salimans2016weight}, in order to guarantee
the continuous flow of information through the network. Other approaches study the properties of the
loss surface and develop training algorithms able to find better minima. In particular, we find
(i) optimization algorithms based on accelerated gradients, like momentum~\cite{polyak1964some,sutskever2013importance}
and Adam~\cite{kingma2014adam}, which combine information about past and current gradients in order 
to dampen the oscillations on the loss surface observed during training, thus converging faster to 
the final solution, and (ii) second-order optimization strategies, like Hessian-free~\cite{martens2010deep} 
and natural gradient methods~\cite{amari1998natural,grosse2015scaling}, 
which looks for efficient approximations of the Hessian using the 
Gauss-Newton~\cite{schraudolph2002fast} and the Fisher information matrices, respectively.
There are a plethora of works studying optimization in neural networks, therefore
we invite the interested reader to see the recent work in~\cite{wilson2017marginal}, 
which provides a theorical comparative analysis between different accelerated gradient-based 
algorithms, and the survey in~\cite{bottou2016optimization}, which presents a more general
overview of optimization strategies in machine learning.

One of the most influential studies about initialization in the last decade is the work 
of~\cite{glorot2010understanding}.
In particular, the authors observes that "the logistic sigmoid activation 
is unsuited for deep networks with random initialization because of its mean value, which can drive 
especially the top hidden layer into saturation". Other more recent works, see for 
example~\cite{xu2016revise}, confirm the fact that the standard logistic activation is more 
difficult to train than other activations and proper rescaling, making the logistic function
similar to the hyperbolic tangent, is required for successful training.
Alternative activations are therefore proposed in the literature. The rectifier linear function
is one of the most appealing solutions~\cite{glorot2011deep},\footnote{Many different extensions
are also proposed. For example,~\cite{he2015delving} parameterize the rectifier linear function
to allow non-zero gradient in the negative region.} because of its unbounded nature that
allow the gradient not to vanish. Principled criteria based on orthogonality~\cite{saxe2013exact} 
and normalization of weights~\cite{mishkin2015all} are used in combination with random
weight initialization to find better starting solutions for training these networks.

Recent theoretical analysis on the properties of random inizialization, 
see~\cite{sussillo2014random} for the analysis of rectifier linear functions 
and~\cite{schoenholz2016deep} for a more general theory validated also on 
hyperbolic tangents, reveals that there exists a range of values for the
variance of weights which are more suited for the propagation of gradients,
thus improving the trainability of networks. Our work proposes to study
the more difficult problem of training standard logistic activations from the 
initialization perspective. Furthermore, we shed light on a more general criterion 
to derive initializating conditions, that explicitly maximizes the amount of information
propagation in neural networks.

%
{\section{Statistical Background}\label{sec:statistical}}
\begin{figure}[!t]
  \centering
      \includegraphics[clip,trim=5cm 12cm 5cm 12cm, width=\linewidth]{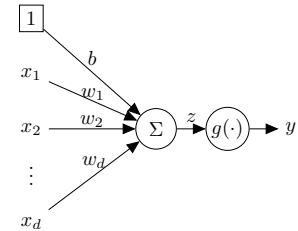}
      \caption{Graphical visualization of a single neuron.}
      \label{fig:neuron}
  \end{figure}
  Let us recall the statistical properties of a single neuron characterized 
  by an input vector $\mathbf{x}\in\mathbb{R}^d$, a weight vector 
  $\mathbf{w}\in\mathbb{R}^d$, a bias $b\in\mathbb{R}$ and a standard logistic activation
  function $g(z)=1/(1+e^{-z})$. Fig.~\ref{fig:neuron} provides a graphical
  interpretation of the computational unit used in many neural networks. 
  
  Consider $z$ as the logit of the given neuron, namely $z=\sum_{i=1}^dw_ix_i+b=\mathbf{w}^T\mathbf{x}+b$.
  By modelling the inputs as independent random variables, with densities/distributions characterized 
  by finite means and finite variances, it is possible to exploit the Lyapunov theorem\footnote{This is an extension of the 
  central limit theorem, which relaxes the assumptions over the random variables and require that
  the random variables are independent but not necessarily identically distributed.} 
  and model $z$ as a Gaussian random variable with mean $\mu$ and variance $\sigma^2$. 
  In this case, $\mu=\sum_{i=1}^dw_iE\{x_i\}+b$ and $\sigma^2$=$\sum_{i=1}^dw_i^2Var\{x_i\}$, where 
  $E\{\cdot\}$ and $Var\{\cdot\}$ are the expected value and the variance operators, respectively.
  It is interesting to note that the mean value of the Gaussian density associated with $z$ is mainly dominated by the 
  parameter $b$ (especially when $E\{x_i\}=0,\forall i =1,\dots,d$, since $\mu=b$), while its variance
  is influenced only by the weight vector $\mathbf{w}$. Therefore, all subsequent considerations valid for $\mu$ and $\sigma$
  will be valid also for $b$ and $\mathbf{w}$, respectively.
  
\begin{figure}[!t]
  \centering
      \subfloat[]{\includegraphics[width=0.5\linewidth]{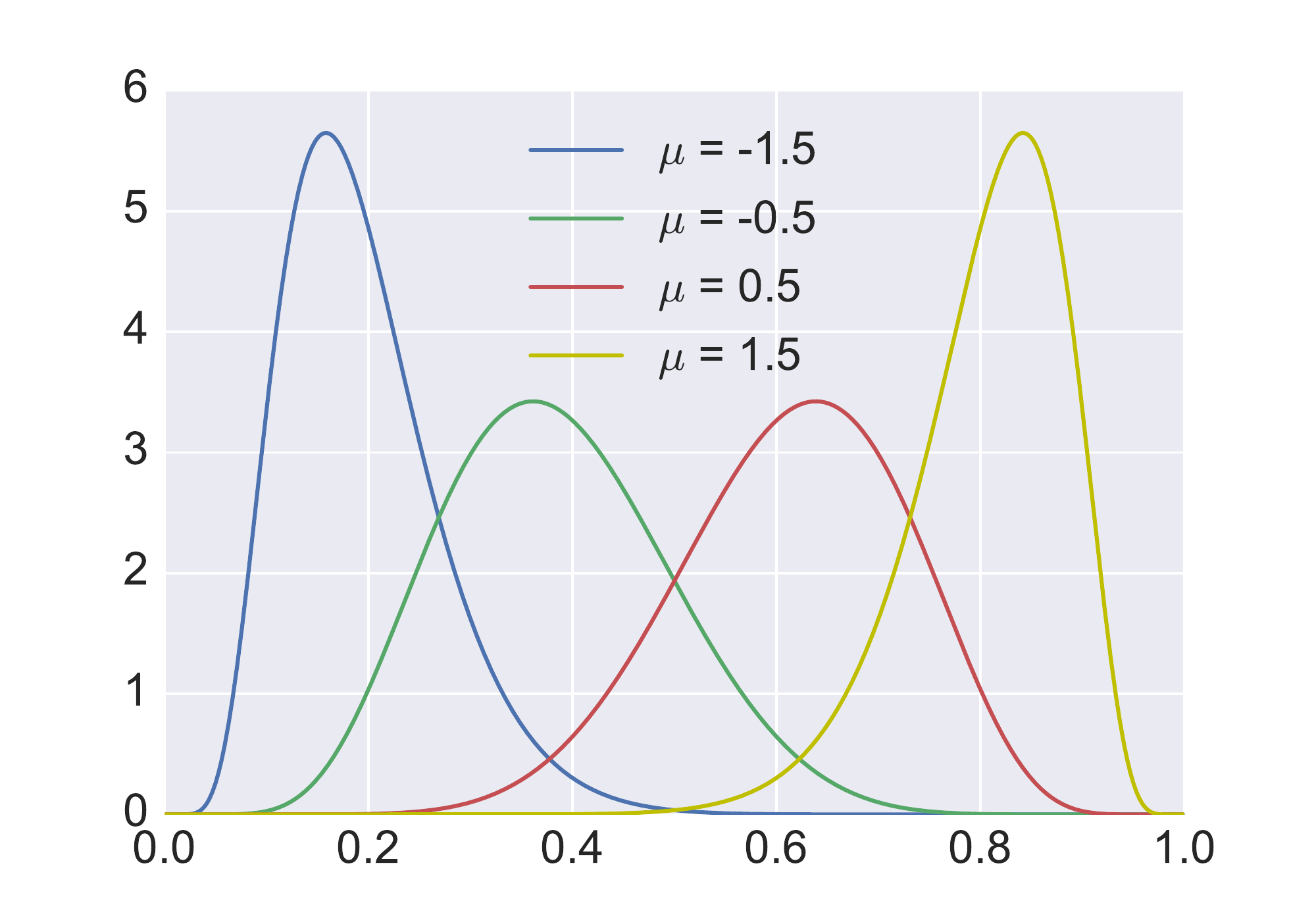}}
      \hfill
      \subfloat[]{\includegraphics[width=0.5\linewidth]{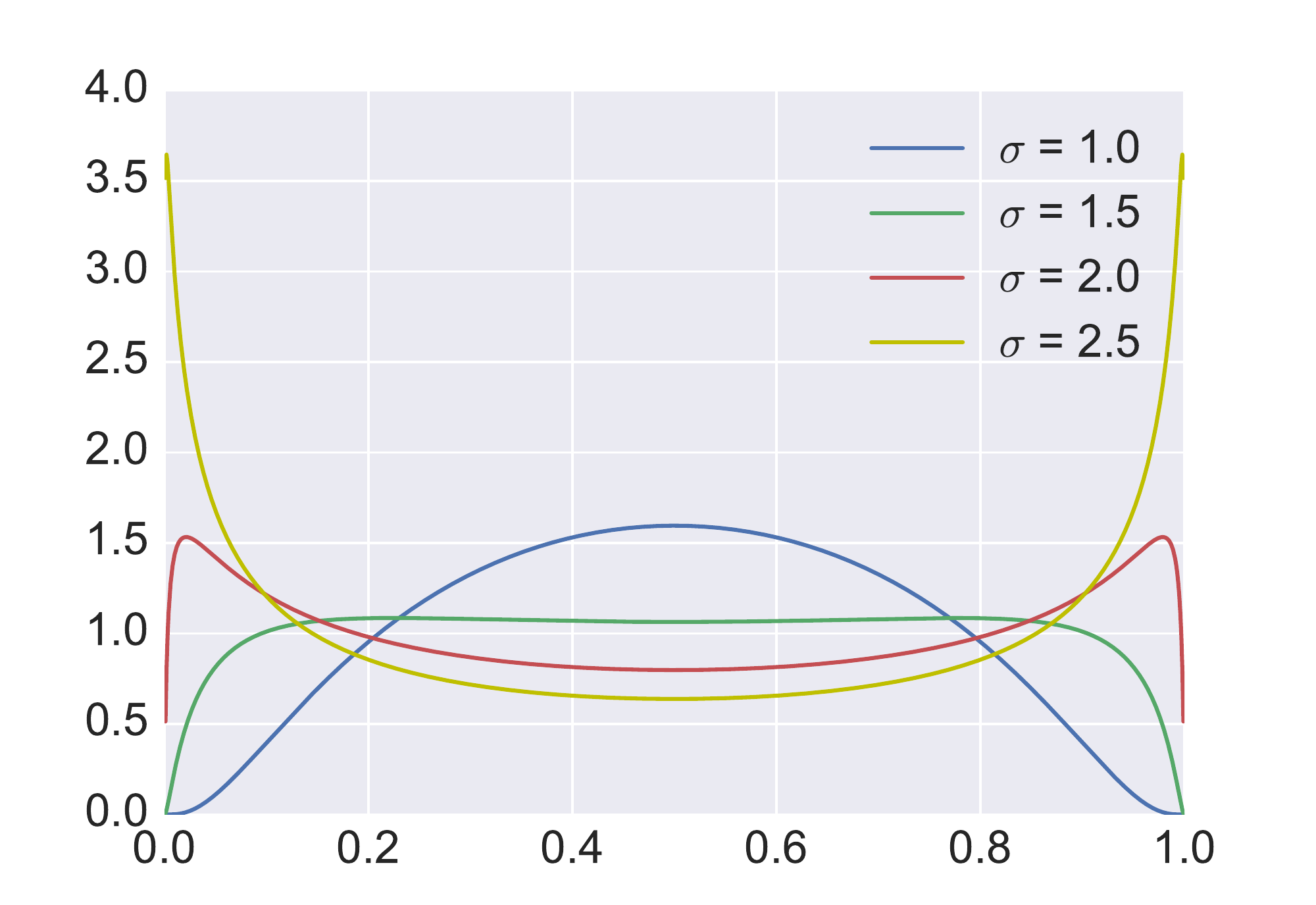}}
      \caption{Visualization of density $p_y(y)$ for different settings of $\mu, \sigma$. 
      (a) $\sigma=0.5$ and variable $\mu$, (b) $\mu = 0$ and variable $\sigma$.}
      \label{fig:density}
  \end{figure}

  Due to its nonlinearity, the activation function $g(z)$ produces an output with different statistical properties from the ones
  associated with logit $z$.
  In fact, the density associated with output $y\in(0,1)$ can be expressed by the following relation, namely:
  \begin{equation}
  \label{eq:density}
  p_y(y)=\frac{1}{y(1-y)\sqrt{2\pi\sigma^2}}\exp\bigg\{-\frac{(\ln\frac{y}{1-y}-\mu)^2}{2\sigma^2}\bigg\}
  \end{equation}
  which is not any more a Gaussian density, as can be seen from Fig.~\ref{fig:density}.
  It is important to mention that $\mu$ and $\sigma$ controls the mean and variance of $p_y(y)$,
  and therefore also the amount of information that can be propagated through the neuron.
  In fact, for very large or very small values of $\mu$, we have no information propagation, since the activation function 
  is saturated and the output variance tends to zero (see Fig.~\ref{fig:density}(a)). 
  The same happens for small values of $\sigma$, because in this case $p_y(y)$ behaves similarly 
  to a Dirac delta centered at $g(\mu)$.
  It can be shown that, when $\sigma\rightarrow\infty$, the density $p_y(y)$ approaches a Bernoulli distribution.\footnote{$p_y(y)\rightarrow 0$ 
  on the interval $(0,1)$ and the mass fully concentrates on the extrema of the interval.} This is
  an extreme case, where information can be propagated, but the ouput becomes discrete (see Fig.~\ref{fig:density}(b)).
  
  In the next section, the behaviour of the neuron is analyzed more in detail from 
  the perspective of information theory.

%
{\section{Information-Theoretic Analysis}\label{sec:information}}
In this section, we ask the following question: which region of the parameter space $(\mu,\sigma)$ 
guarantees that the maximum amount of information is propagated through the activation function 
$g(z)$?
We address this question by formulating the problem as an optimization.

Logit $z$ and output $y$ are modelled as continuous random variables
distributed according to $\mathcal{N}(\mu,\sigma^2)$ and $p_y(y)$, respectively.
We choose the entropy of $y$, viz. $H(y)$, as the objective of the maximization
problem and discard for example the mutual information, since it is not defined 
for this particular case.\footnote{It is not defined because of
Dirac delta distributions, which come from the fact that $g(\cdot)$ is deterministic
and that the variables are continuous.}
Therefore, the objective can be written in the following way:
\begin{IEEEeqnarray}{lll}
    \label{eq:entropy}
    H(y) &\doteq & -\int_{0}^{1}p_y(y)\ln p_y(y)dy \nonumber\\
         &=& \int_{-\infty}^{\infty}\frac{e^{{-}\frac{(z-\mu)^2}{2\sigma^2}}}{\sqrt{2\pi\sigma^2}}
         \Bigg\{\frac{(z-\mu)^2}{2\sigma^2}+ \nonumber\\
         & & +\ln\Big[g(z)\big(1-g(z)\big)\sqrt{2\pi\sigma^2}\Big]\Bigg\}dz \nonumber\\
         &=& \int_{-\infty}^{\infty}\frac{e^{{-}\frac{(z-\mu)^2}{2\sigma^2}}}{\sqrt{2\pi\sigma^2}}
         \Bigg\{\frac{(z-\mu)^2}{2\sigma^2}+\ln\sqrt{2\pi\sigma^2}+\ln g'(z)\Bigg\}dz \nonumber\\
         &=& H(z)+E_z\{\ln g'(z)\}
\end{IEEEeqnarray}
where $H(z)$, $E_z\{\cdot\}$, $g'(z)$ are the entropy, the expected
value and the derivative of $g(\cdot)$ computed on variable $z$, respectively.
Note that the second line in~(\ref{eq:entropy}) is obtained from the first one by a simple 
change of variable, namely $y=g(z)$. Thus, the information coming out 
from the neuron is proportial to the information of the logit and the shape of 
the activation function. 

\begin{figure}[!t]
\centering
    \includegraphics[clip, trim=2cm 1.3cm 2cm 1.3cm, width=0.7\linewidth]{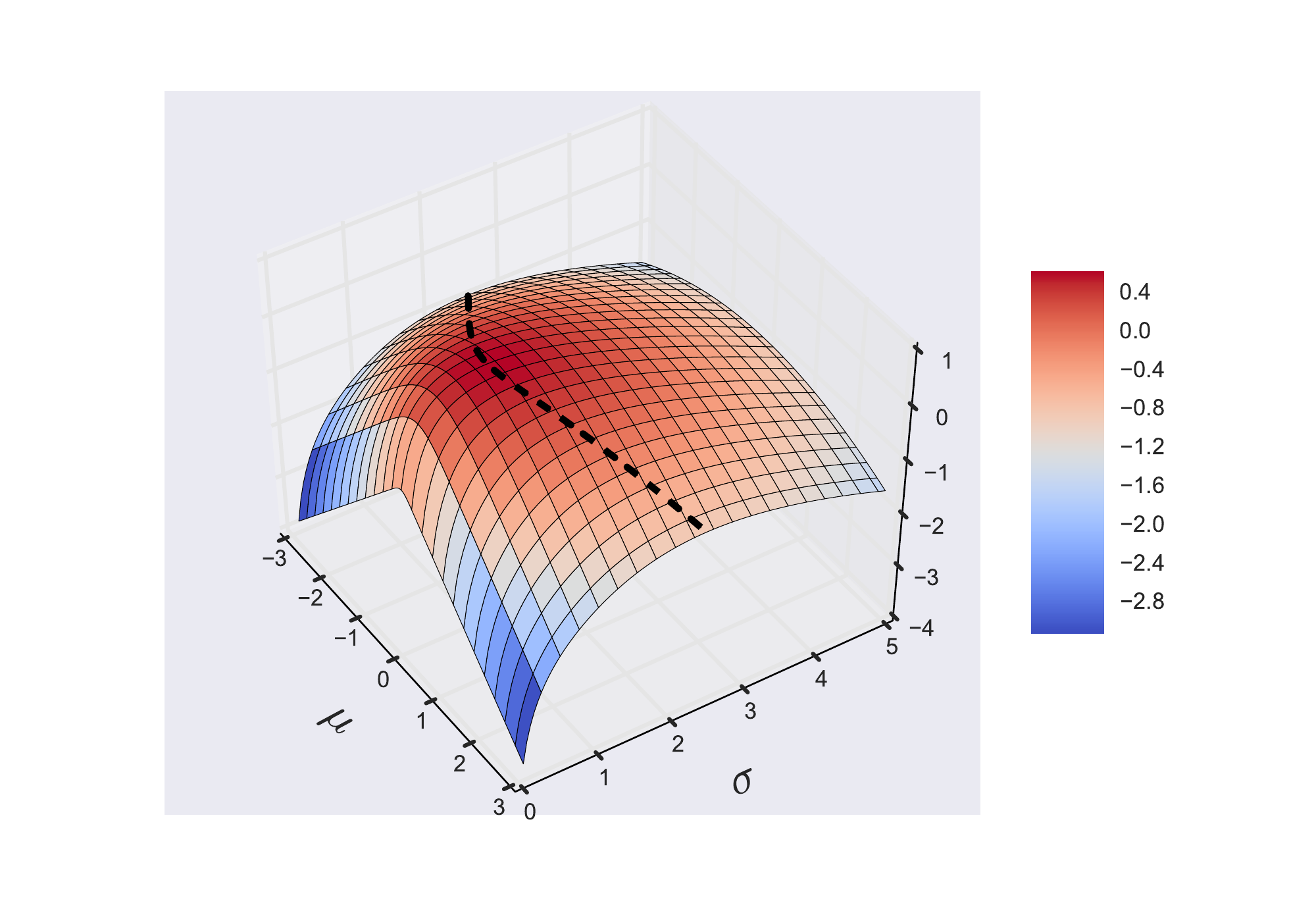}
    \caption{3D plot of $H_{B}(\mu,\sigma)$. The dotted curve on the surface represents the set of points
    for which $\frac{\partial H_{B}(\mu,\sigma)}{\partial\sigma}=0$.}
    \label{fig:lower}
\end{figure}

In general, the term $E_z\{\ln g'(z)\}$ in~(\ref{eq:entropy})
is difficult to compute analytically and can be approximated using a lower and an upper 
bound (see Appendix~\ref{sec:details}). This brings us to the following inequalities:
\begin{equation}
    \label{eq:bounds}
    H_B(\mu,\sigma) -2\ln 2 \leq H(y) < H_B(\mu,\sigma)
\end{equation}
and
\begin{IEEEeqnarray}{ll}
    \label{eq:lowerbound}
    H_B(\mu,\sigma)\doteq& \frac{1}{2} + \frac{1}{2}\ln(2\pi\sigma^2) - \nonumber\\
                    & -\mu\:erf\bigg(\frac{\mu}{\sigma\sqrt{2}}\bigg)
                      - \frac{2\sigma}{\sqrt{2\pi}}e^{-\frac{\mu^2}{2\sigma^2}}
\end{IEEEeqnarray}
where $erf(\cdot)$ is the error function~\cite{papoulis2002probability} and the 
function $H_B(\mu,\sigma)$ can be visualized in Fig.~\ref{fig:lower}.
From~\ref{eq:bounds}, $H_{B}(\mu,\sigma)$ defines both the lower
and upper bounds of $H(y)$. Therefore, it can be used as a surrogate objective for our maximization problem 
to find the optimal values of $\mu$ and $\sigma$. 
Based on this principle, we can enunciate the following theorem (proof in Appendix~\ref{sec:proof}).
\begin{theorem}
\label{th:stationarity}
The optimality conditions for $H_{B}(\mu,\sigma)$, defined in~(\ref{eq:lowerbound}),
are given by the following statements:
\begin{itemize}
    \item $\forall\sigma\in\mathbb{R}^+$, $\frac{\partial H_{B}(\mu,\sigma)}{\partial\mu}=0\:\Rightarrow\:\mu=0$,
    \item $\forall\mu\in\mathbb{R}$, $\frac{\partial H_{B}(\mu,\sigma)}{\partial\sigma}=0\:\Rightarrow\:
           \sigma=\sqrt{\frac{\pi}{2}}e^{\frac{W\big(\frac{2\mu^2}{\pi}\big)}{2}}$,
\end{itemize}
\par where $W(\cdot)$ is the principal branch of the Lambert W function~\cite{corless1996lambertw}.
The optimal solution, maximizing $H_{B}(\mu,\sigma)$, is therefore obtained at the point $\mu=0$ and
$\sigma=\sqrt{\frac{\pi}{2}}\approx 1.2533$.
\end{theorem}
The optimality result given by Theorem~\ref{th:stationarity} is interpreted as a condition
on $\mu$ and $\sigma$ to have maximum amount of information propagated through the sigmoid
activation. On one hand, the condition $\mu=0$ implies that $b=-\sum_{i=1}^dw_iE\{x_i\}$ (following
directly from Section~\ref{sec:statistical}), meaning that the expected value associated with 
the logit $z$ must lie in the central part of the sigmoid far away from its saturating regions. 
On the other hand, the condition $\sigma=\sqrt{\pi/2}$ for $\mu=0$ implies that $\sum_{i=1}^dw_i^2Var\{x_i\}=\pi/2$
(following directly from Section~\ref{sec:statistical}). Note that, if the variances of 
inputs are equal to the same constant, namely $Var\{x_i\}=k$ for all 
$i=1,\dots,d$,\footnote{This is a common assumption used in the theory of neural 
networks~\cite{lecun2012efficient}.} 
then $\|\mathbf{w}\|_2=\sqrt{\pi/(2k)}$. This means that
the maximum amount of information propagated thorugh $g(z)$ is obtained when vector 
$\mathbf{w}$ lies on the hypersphere of radius $\sqrt{\pi/(2k)}$. In Appendix~\ref{sec:general},
we show how to deal with the more general case where the input variances are different
from each other. In practice, condition $\sum_{i=1}^dw_i^2Var\{x_i\}=\pi/2$ may be rewritten as
$\mathbf{w}^T\mathbf{D}\mathbf{w}=\pi/2$, where $\mathbf{D}\doteq diag(Var\{x_1\},
\dots,Var\{x_d\})$. If $\mathbf{D}$ is positive definite, then the quadratic equation 
$\mathbf{w}^T\mathbf{D}\mathbf{w}=\pi/2$ characterizes a multidimensional ellipsoid. 
Vector $\mathbf{w}$ must therefore lie on this geometric locus of points to guarantee
the maximal information propagation.

We argue that the obtained results are useful to define the initial conditions for
learning the parameters of neurons with standard logistic activations and that cannot
be enforced during training, because they limit the expressivity of the neural network. 
In fact, they impose that each neuron lies in the linear region of its sigmoid activation,
thus constraining the whole network to approximate only a linear function.

In the next sections, we provide theoretical evidence about the utility of these conditions.

%
{\section{The Problem of Vanishing Gradients}\label{sec:vanishing}}
In this section, we study the implications of maximizing the mutual
information at each neuron on the problem of vanishing gradients~\cite{bengio1994learning,
pascanu2013difficulty}. In particular, we will show that the conditions established by 
Theorem~\ref{th:stationarity} ensure that the selected starting point lies far away from 
the critical point that implies the occurrence of the vanishing gradients.
Although the proposed theoretical analysis does not imply that this condition cannot be 
reached at some later stage of the learning, the effectiveness of the proposed initialization 
will be also supported by the extensive testing, presented in the experimental section. 

We study the problem of vanishing gradients by adopting the same methodological 
analysis of~\cite{bengio1994learning,pascanu2013difficulty} and focus on recurrent 
neural networks, which can be seen as the deepest version of feedforward neural 
nets. The obtained results are therefore valid for traditional feedforward 
neural networks~\cite{bengio1994learning,pascanu2013difficulty}.

Recall that a recurrent neural network is fully described by the following
equations:
\begin{IEEEeqnarray}{ll}
\label{eq:recurrence}
    \mathbf{\mathcal{E}}^t &= g(\mathbf{W}_{out}\mathbf{y}^t+\mathbf{b}_{out}) \nonumber\\
    \mathbf{y}^t &= g(\mathbf{W}\mathbf{y}^{t-1}+\mathbf{W}_{in}\mathbf{x}^t+\mathbf{b})
\end{IEEEeqnarray}
where $t$ is used to identify time, or equivalently to indicate the layer in a 
deep feedforward neural net. $\mathbf{x}^t\in\mathbb{R}^d$, $\mathbf{y}^t
\in\mathbb{R}^h$ and $\mathbf{\mathcal{E}}^t\in\mathbb{R}^o$ are the input, 
the hidden state and the output vectors of the network, respectively.
$\mathbf{W}_{in}\in\mathbb{R}^{h\times d}$, $\mathbf{W}\in\mathbb{R}^{h\times h}$ and $\mathbf{W}_{out}\in
\mathbb{R}^{o\times h}$ are the weight matrices associated with the connections of 
neurons, $\mathbf{b}\in\mathbb{R}^h$ and $\mathbf{b}_{out}\in\mathbb{R}^o$ are
the bias vectors and $g(\cdot)$ is an element-wise operator that
applies the sigmoid function to the incoming vector.

Recurrent neural networks are usually trained by minimizing an objective 
$\mathcal{L}$ that sums all loss contributions incurred over a time horizon 
of duration $T$, namely $\mathcal{L}=\sum_{t=1}^T\mathcal{L}^t$, where 
$\mathcal{L}^t:\mathbb{R}^o\rightarrow\mathbb{R}$. The training requires
computing the gradient of the objective with respect to parameters $\mathbf{W}$~
\cite{rumelhart1986learning}, namely:
\begin{IEEEeqnarray}{ll}
\label{eq:gradients}
    \frac{\partial\mathcal{L}}{\partial\mathbf{W}} &=
    \sum_{t=1}^{T}\frac{\partial\mathcal{L}^t}{\partial\mathbf{W}} \nonumber\\
    \frac{\partial\mathcal{L}^t}{\partial\mathbf{W}} &=
    \sum_{k=1}^{T}diag\bigg(\frac{\partial\mathcal{L}^t}{\partial\mathbf{y}^k}\bigg)
    \frac{\partial\mathbf{y}^k}{\partial\mathbf{W}} \nonumber\\
    \frac{\partial\mathcal{L}^t}{\partial\mathbf{y}^k} &=
    \bigg(\frac{\partial\mathcal{L}^t}{\partial\mathbf{\mathcal{E}}^t}^T
    \frac{\partial\mathbf{\mathcal{E}}^t}{\partial\mathbf{y}^t}
    \frac{\partial\mathbf{y}^t}{\partial\mathbf{y}^k}\bigg)^T
    \nonumber\\
    \frac{\partial\mathbf{y}^t}{\partial\mathbf{y}^k} &=
    \prod_{l=t}^{k+1}\frac{\partial\mathbf{y}^l}{\partial\mathbf{y}^{l-1}} = 
    \prod_{l=t}^{k+1}diag\big(g'(\mathbf{z}^l)\big)\mathbf{W}
\end{IEEEeqnarray}
where $\mathbf{z}^l=\mathbf{W}\mathbf{y}^{l-1}+\mathbf{W}_{in}\mathbf{x}^l+\mathbf{b}$
is the vector of logits.

By exploiting the fact that for any real matrices $\mathbf{A}$,
$\mathbf{B}$ and for any real vector $\mathbf{v}$, $\|\mathbf{A}\mathbf{v}\|_2
\leq\|\mathbf{A}\|_*\|\mathbf{v}\|_2$ and $\|\mathbf{A}\mathbf{B}\|_*
\leq\|\mathbf{A}\|_*\|\mathbf{B}\|_*$, where $\|\cdot\|_*$ is the 
operator-2 norm, it is possible to derive an upper-bound on the gradient norm
$\|\frac{\partial\mathcal{L}^t}{\partial\mathbf{y}^k}\|_2$. In fact,
\begin{IEEEeqnarray}{ll}
    \label{eq:upperbound}
    \bigg\|\frac{\partial\mathcal{L}^t}{\partial\mathbf{y}^k}\bigg\|_2 &=
    \bigg\|\frac{\partial\mathcal{L}^t}{\partial\mathbf{\mathcal{E}}^t}^T
    \frac{\partial\mathbf{\mathcal{E}}^t}{\partial\mathbf{y}^t}
    \frac{\partial\mathbf{y}^t}{\partial\mathbf{y}^k}\bigg\|_2 \nonumber\\
    &\leq \bigg\|\frac{\partial\mathcal{L}^t}{\partial\mathbf{\mathcal{E}}^t}\bigg\|_2
    \bigg\|\frac{\partial\mathbf{\mathcal{E}}^t}{\partial\mathbf{y}^t}\bigg\|_*
    \bigg\|\frac{\partial\mathbf{y}^t}{\partial\mathbf{y}^k}\bigg\|_* \nonumber\\
    &\leq \bigg\|\frac{\partial\mathcal{L}^t}{\partial\mathbf{\mathcal{E}}^t}\bigg\|_2
    \bigg\|\frac{\partial\mathbf{\mathcal{E}}^t}{\partial\mathbf{y}^t}\bigg\|_*
    \prod_{l=t}^{k+1}\bigg\|\frac{\partial\mathbf{y}^l}{\partial\mathbf{y}^{l-1}}\bigg\|_*
\end{IEEEeqnarray}
Note that from~(\ref{eq:gradients}) and~(\ref{eq:upperbound}), we can derive
the following relations:
\begin{IEEEeqnarray}{ll}
    \label{eq:upperbound2}
    \bigg\|\frac{\partial\mathbf{y}^l}{\partial\mathbf{y}^{l-1}}\bigg\|_* &\leq
    \big\|diag\big(g'(\mathbf{z}^l)\big)\big\|_*\big\|\mathbf{W}\big\|_F \nonumber\\
    &= \max_{i=1,\dots,h}\big\{g'(z_i^l)\big\}\big\|\mathbf{W}\big\|_F \nonumber \\
    &\leq \frac{\|\mathbf{W}\big\|_F}{4}
\end{IEEEeqnarray}
In particular, the first inequality in~(\ref{eq:upperbound2}) can be obtained 
using the property $\|\mathbf{A}\|_*\leq\|\mathbf{A}\|_F$, where $\|\cdot\|_F$ 
is the Frobenius norm. The equality in second line follows directly 
from the fact that the operator-2 norm of a diagonal matrix is equal to the 
maximum of its diagonal entries, whereas the last inequality is due to the 
fact that the derivative of a sigmoid function cannot be larger than $1/4$.

By using the result in~(\ref{eq:upperbound2}), the gradient in~(\ref{eq:upperbound})
is bounded by the follwoing inequality:
\begin{IEEEeqnarray}{ll}
    \label{eq:upperbound3}
    \bigg\|\frac{\partial\mathcal{L}^t}{\partial\mathbf{y}^k}\bigg\|_2 &\leq
    \bigg\|\frac{\partial\mathcal{L}^t}{\partial\mathbf{\mathcal{E}}^t}\bigg\|_2
    \bigg\|\frac{\partial\mathbf{\mathcal{E}}^t}{\partial\mathbf{y}^t}\bigg\|_*
    \prod_{l=t}^{k+1}\frac{\|\mathbf{W}\big\|_F}{4} \nonumber\\
    &= \bigg\|\frac{\partial\mathcal{L}^t}{\partial\mathbf{\mathcal{E}}^t}\bigg\|_2
    \bigg\|\frac{\partial\mathbf{\mathcal{E}}^t}{\partial\mathbf{y}^t}\bigg\|_*
    \bigg(\frac{\|\mathbf{W}\big\|_F}{4}\bigg)^{t-k-1}
\end{IEEEeqnarray}
The vanishing gradient problem refers to the decay of $\|\frac{\partial\mathcal{L}^t}
{\partial\mathbf{y}^k}\|_2$ as the number of time instants $t-k$, or equivalently the 
number of layers, becomes larger. A sufficient condition for the occurrence of 
this problem is given by the condition $\|\mathbf{W}\big\|_F<4$, due to the fact
that the bound in~(\ref{eq:upperbound3}) tends to zero as $t-k\rightarrow\infty$.

Note that $\|\mathbf{W}\|_F=\sqrt{\sum_{i=1}^h\|\mathbf{w}_i\|_2^2}$, where $\mathbf{w}_i$
is the $i$-th row of matrix $\mathbf{W}$ corresponding to the weights of the $i$-th neuron.
Furthermore, by the conditions derived in Section~\ref{sec:information}, viz.
$\|\mathbf{w}_i\|_2^2=\pi/(2k)$ for all $i=1,\dots,h$, we have that
\begin{equation}
    \label{eq:condition}
    \|\mathbf{W}\|_F = \sqrt{\frac{\pi h}{2k}}
\end{equation}
If the number of hidden neurons is less than the quantity 
$h_{critic}\doteq32k/\pi$, which implies that~(\ref{eq:condition})
is less than $4$, then the vanishing gradient problem 
is guaranteed to occur. In this case, $k$ refers to the output 
variance of a sigmoid activation and is less than or equal to $1/4$ 
(derived from the fact that the output density of a neuron is limited 
by a Bernoulli distribution and its maximum variance is $1/4$, see 
Section~\ref{sec:statistical}). Therefore, $h_{critic}\leq 8/
\pi\approx2.55$. 

This means that the sufficient condition for
the occurrence of the vanishing gradient problem is met
only when the number of hidden neurons is less than 3. In practical
cases, this limit is overcome with a very large margin, given
that real-world applications usually require hundreds/thousands 
of hidden neurons per layer. Therefore, even if we cannot 
conclude that the whole training is exempt of vanishing gradients
problems, we inizialize the process with a sufficient margin to 
prevent the problem in the initial phase, which is typically the
most critical. Experimental results will provide evidence of this,
as shown later in the paper.

%
{\section{Experimental Results}\label{sec:experiments}}
In this section, we evaluate the performance of the proposed initialization theory on several benchmarks.
We start by analyzing shallow networks, then consider the case of deep networks and finally extend the
analysis to recurrent neural networks (RNNs, being even deeper than previous cases). We compare our initialization against several
competitors. Hereunder, we provide a summary of all strategies:
\begin{itemize}
    \item Lecun et al.~\cite{lecun2012efficient} initialize the weights according to 
    $w_{ij}^l\sim U\bigg(-\frac{1}{\sqrt{n^l}},\frac{1}{\sqrt{n^l}}\bigg)$, 
    where $w_{ij}^l$ is the weight connecting neuron $i$ with neuron $j$ in layer $l$ and $n^l$ is the 
    number of input neurons to layer $l$. The biases are set to zero.

    \item Glorot and Bengio~\cite{glorot2010understanding} (also known as "Xavier initialization") 
    initialize the weights according to $w_{ij}^l\sim U\bigg(-\frac{\sqrt{6}}{\sqrt{n^l+n^{l+1}}},\frac{\sqrt{6}}{\sqrt{n^l+n^{l+1}}}\bigg)$. 
    The biases are set to zero.
    
    \item Saxe et al.~\cite{saxe2013exact} propose an initialization where each weight matrix 
    is randomly generated from the family of orthogonal matrices, 
    namely $\mathbf{W}^l$ is satisfying the relation ${\mathbf{W}^l}^T\mathbf{W}^l=\mathbf{I}$. 
    The biases are set to zero.
    
    \item Mishkin and Matas~\cite{mishkin2015all} extend the initialization of Saxe et al. by 
    normalizing the weight matrix by the output variance in each layer. The biases are set to zero.
    
    \item Our approach exploits the result of Theorem~\ref{th:stationarity} in Section~\ref{sec:information}, 
    and initializes the weights according to
    $w_{ij}^l=\sqrt{\frac{\pi}{2k}}\frac{\tilde{w}_{ij}^l}{\|\tilde{\mathbf{w}}_j^l\|_2}$,\footnote{$k\doteq Var\{x\}\approx0.0589$, 
    where $Var\{x\}$ is computed numerically using $p_y(y)$ at optimality ($x$ is therefore the output of the 
    previous neuron).} 
    where $\tilde{w}_{ij}^l$ can be obtained by using either random generation, namely $\tilde{w}_{ij}^l\sim U(-1,1)$,
    or the orthogonal initialization in~\cite{saxe2013exact}. The 
    biases are set according to $b_j^l=-\sum_{i=1}^{n^l}w_{ij}^l E\{x_i^l\}$, 
    where $E\{x_i^l\}$ is the expected value of input neuron $x_i^l$.
\end{itemize}
In the experiments, we use \textit{lecun}, \textit{glorot}, \textit{ortho} and \textit{lsuv} to identify 
the results achieved by~\cite{lecun2012efficient},\cite{glorot2010understanding},\cite{saxe2013exact} and~\cite{mishkin2015all}, 
respectively. \textit{random+EP} and \textit{ortho+EP} are instead used to identify the two versions
of our initialization procedure. In this case, the acronym \textit{EP} stands for \textit{ellyptical projection} 
(see Section~\ref{sec:information} for a detailed discussion). 

All experiments presented in the next sections are run on a Linux machine equipped by 4 cpu @ 3.2 GHz, 
16 GB RAM and a GPU card (NVIDIA TITAN X).


\subsection{XOR Case: Shallow Network}
\begin{figure*}[!t]
\centering
    \subfloat[]{\includegraphics[width=0.3\textwidth]{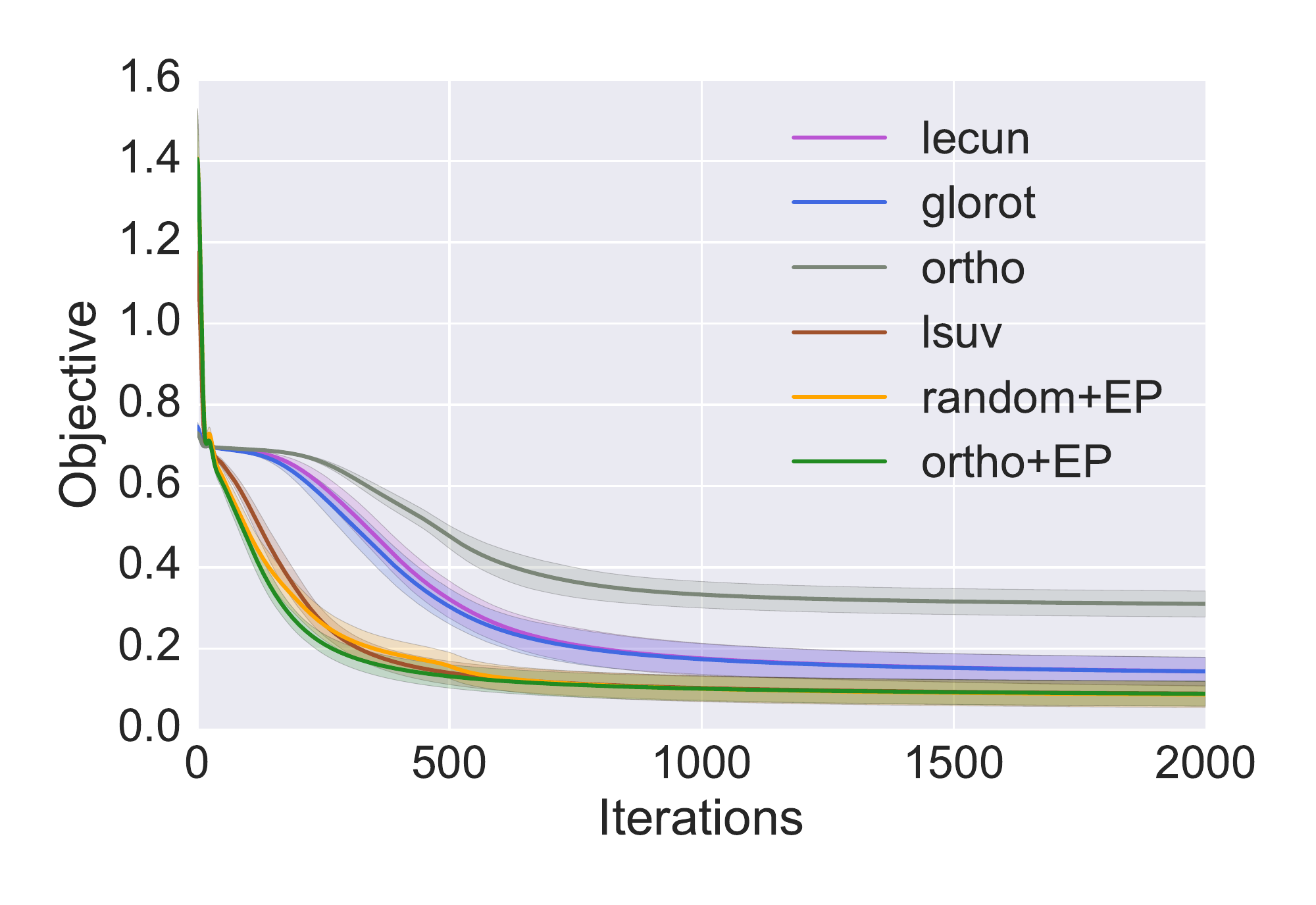}}
    \hfil
    \subfloat[]{\includegraphics[width=0.3\textwidth]{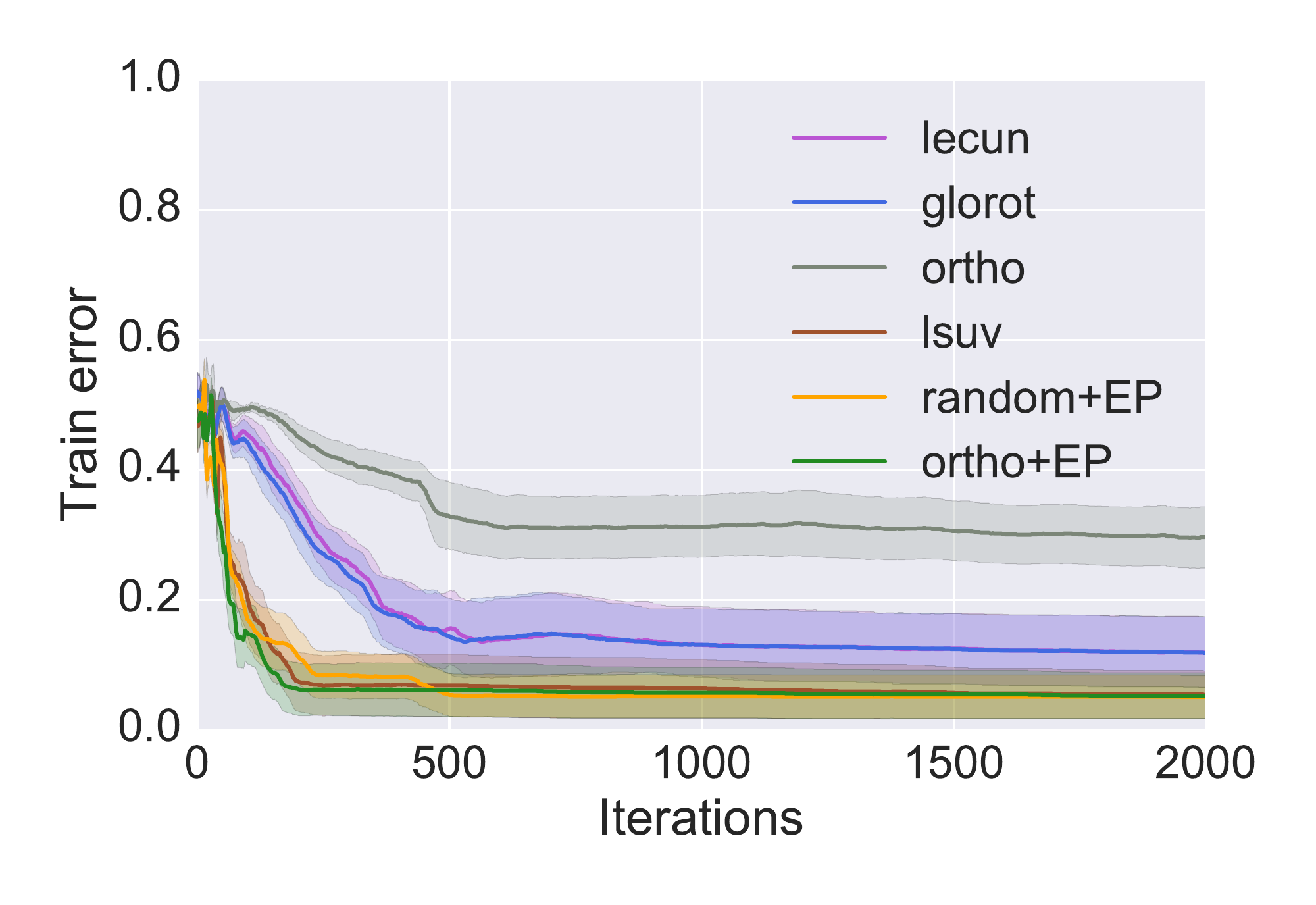}}
    \hfil
    \subfloat[]{\includegraphics[width=0.3\textwidth]{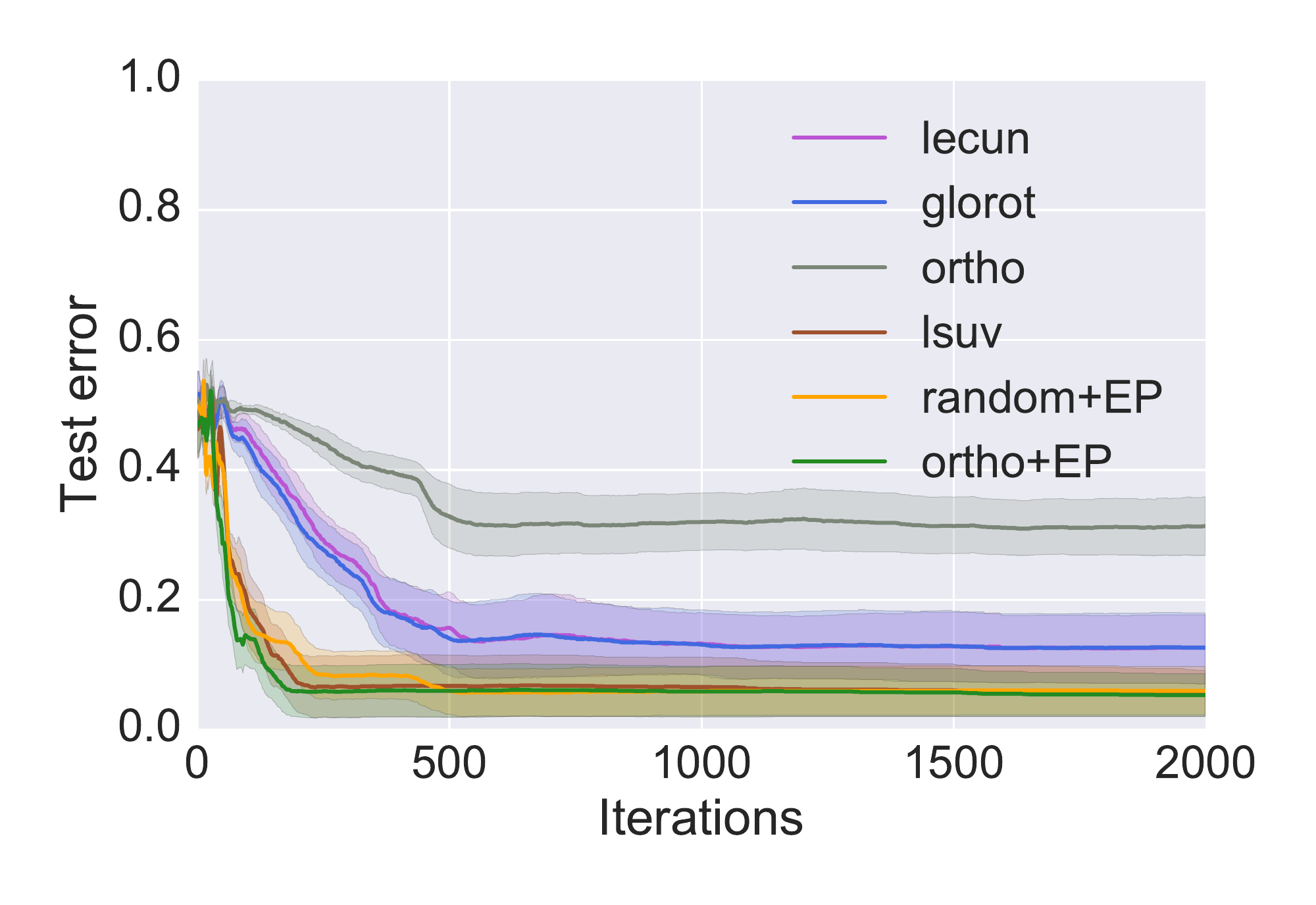}}
    \caption{Learning curves for (a) training objective, (b) training error and (c) test error on the XOR case over different initialization methods.}
    \label{fig:xor}
\end{figure*}
In this experiment, we generate 200 samples from four 2-dimensional Gaussians. Classes are assigned to 
create a XOR configuration. In this case, we use a neural network with one hidden layer containing two 
hidden units. This is the smallest network that can learn correctly the problem (4 modes can be represented 
more compactly with 2 binary digits, namely 2 hidden neurons). Gradient descent with momentum 
equal to $0.9$ and learning rate equal to $0.05$ is applied to the minimization of the cross-entropy 
objective function. Fig.~\ref{fig:xor} shows results over ten repeated experiments 
(generating new data from the same distribution and initializing the network differently). It is clear from these experiments that our proposed initialization allows to achieve the best solution 
in a very efficient way. Also, \textit{lsuv} is able to achieve comparable performance to our method. Moreover, it's
interesting to note that \textit{ortho} obtains the worst performance, in fact the test error rate, viz. T.E., is about 25\%, meaning
that this initialization is not particularly suited for this non-linear separable scenario.

\subsection{MNIST: Shallow Network}
\begin{figure*}[!t]
\centering
    \subfloat[]{\includegraphics[width=0.3\textwidth]{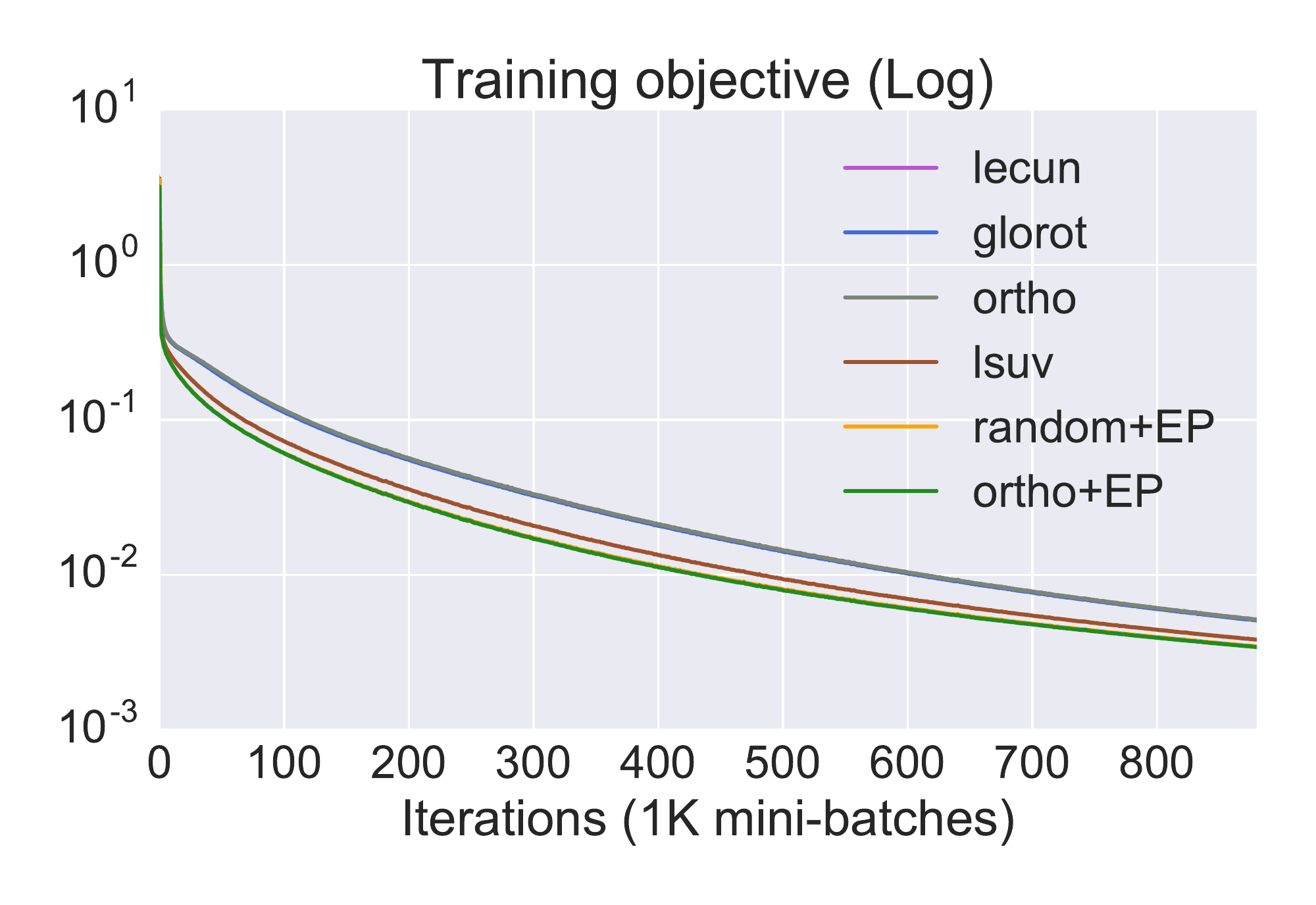}}
    \hfil
    \subfloat[]{\includegraphics[width=0.3\textwidth]{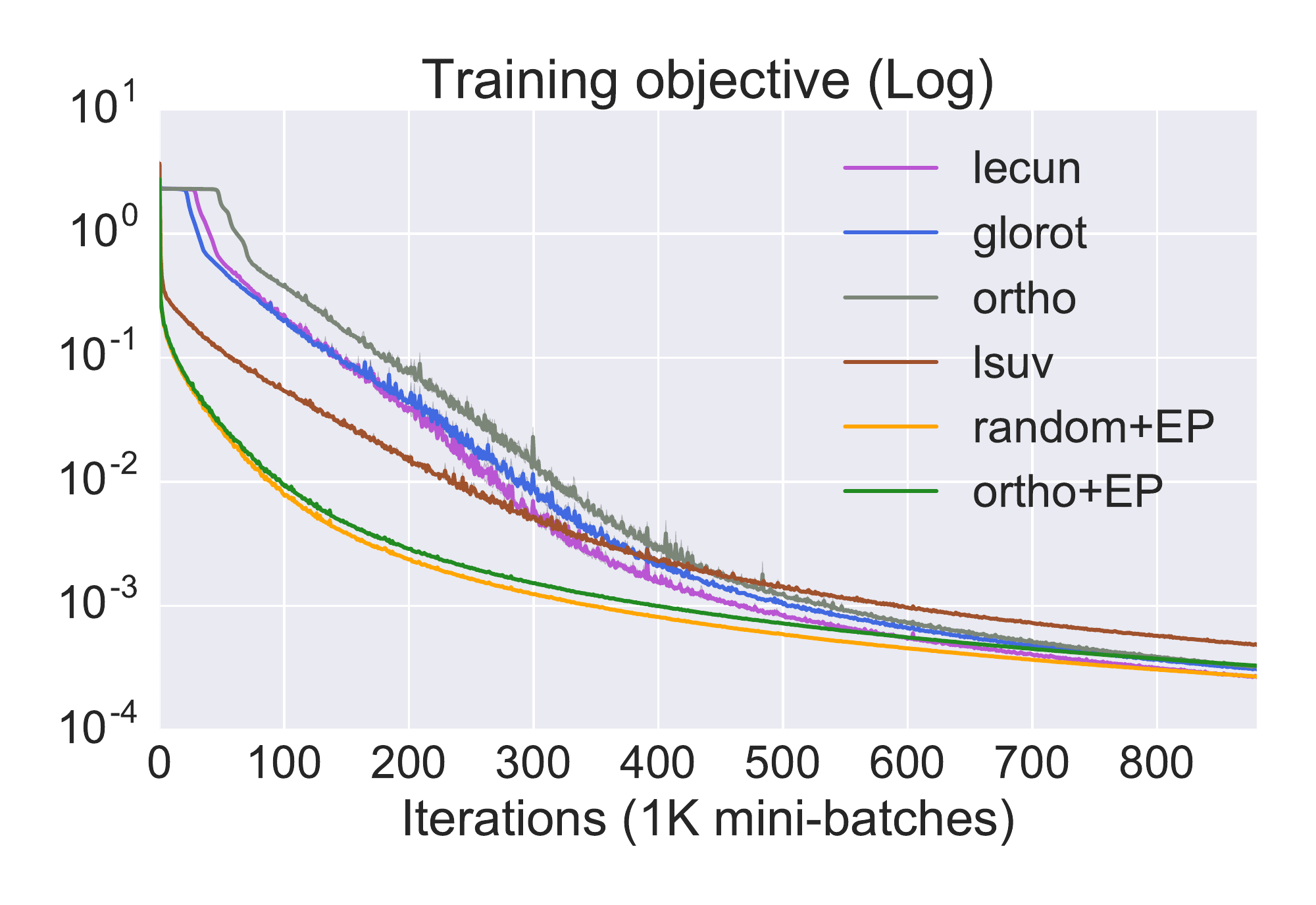}}
    \hfil
    \subfloat[]{\includegraphics[width=0.3\textwidth]{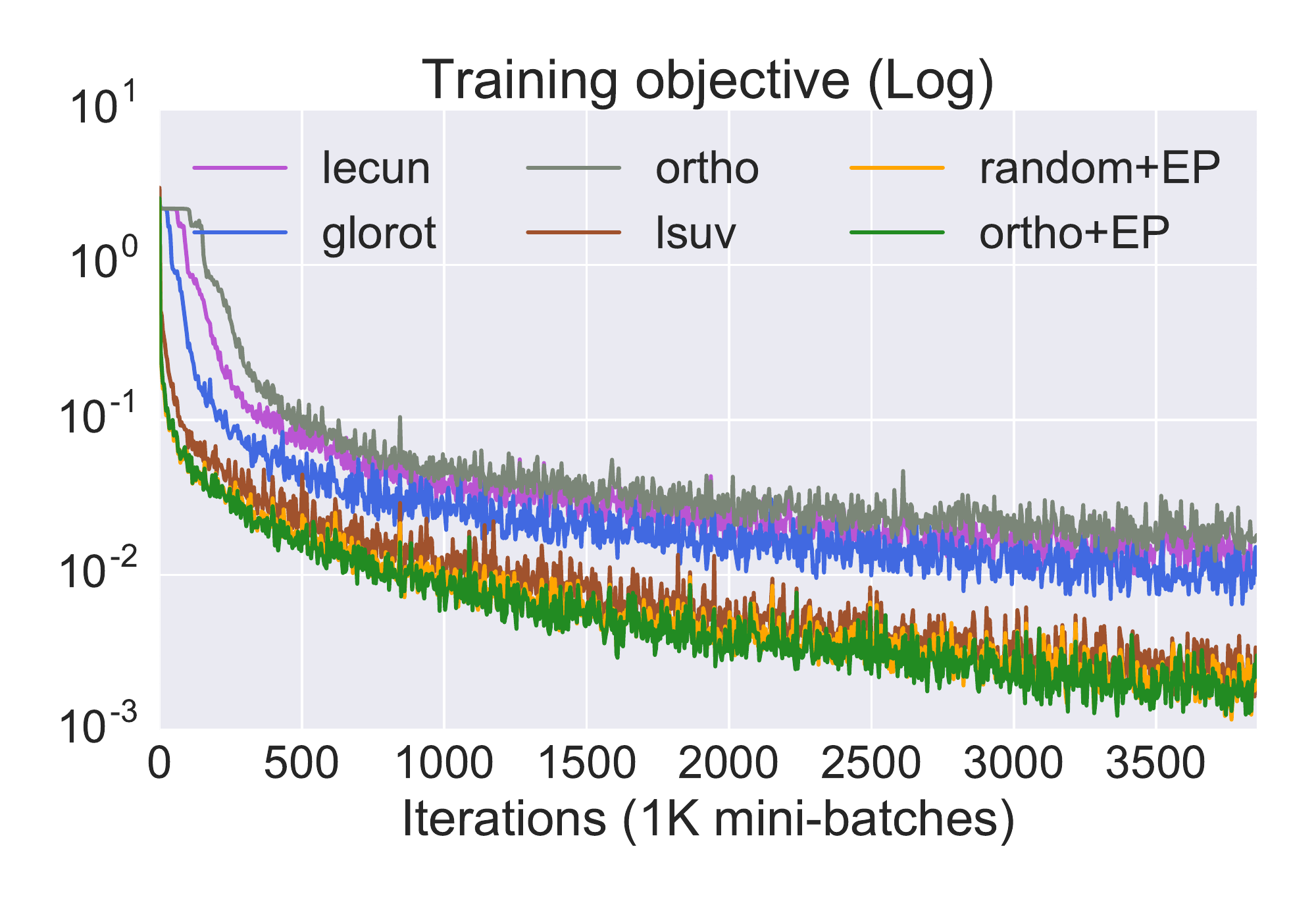}}
    \hfil
    \subfloat[]{\includegraphics[width=0.3\textwidth]{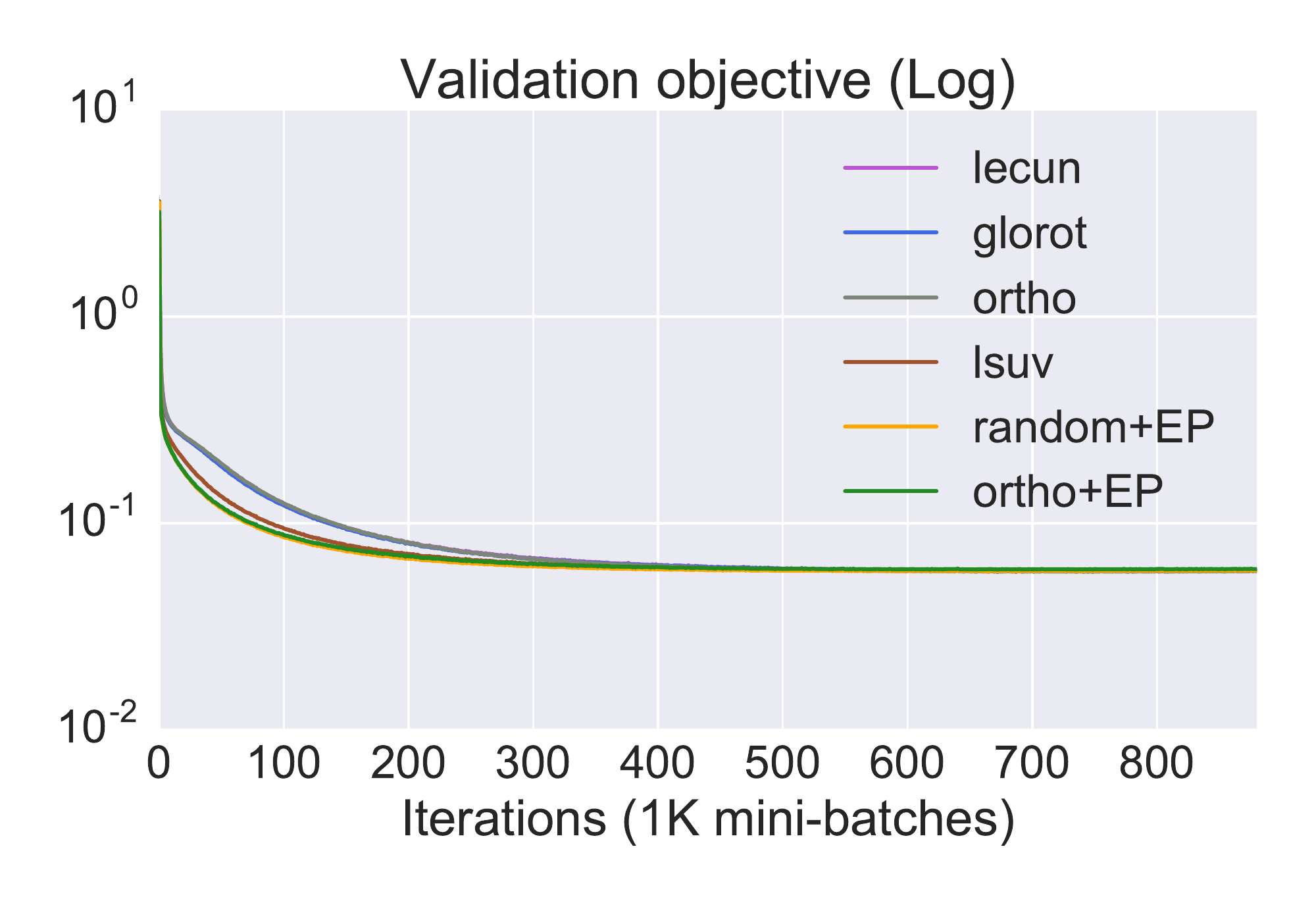}}
    \hfil
    \subfloat[]{\includegraphics[width=0.3\textwidth]{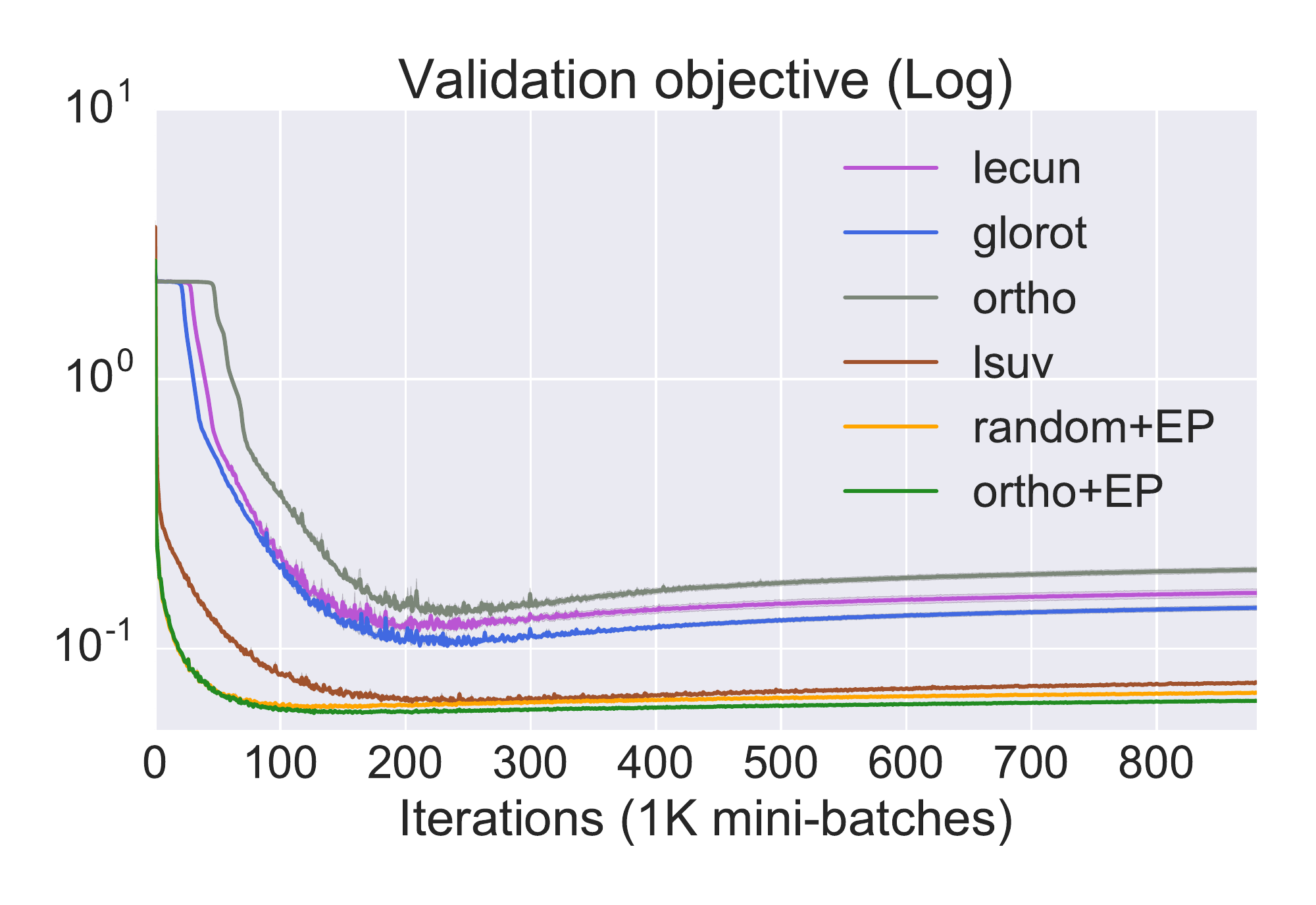}}
    \hfil
    \subfloat[]{\includegraphics[width=0.3\textwidth]{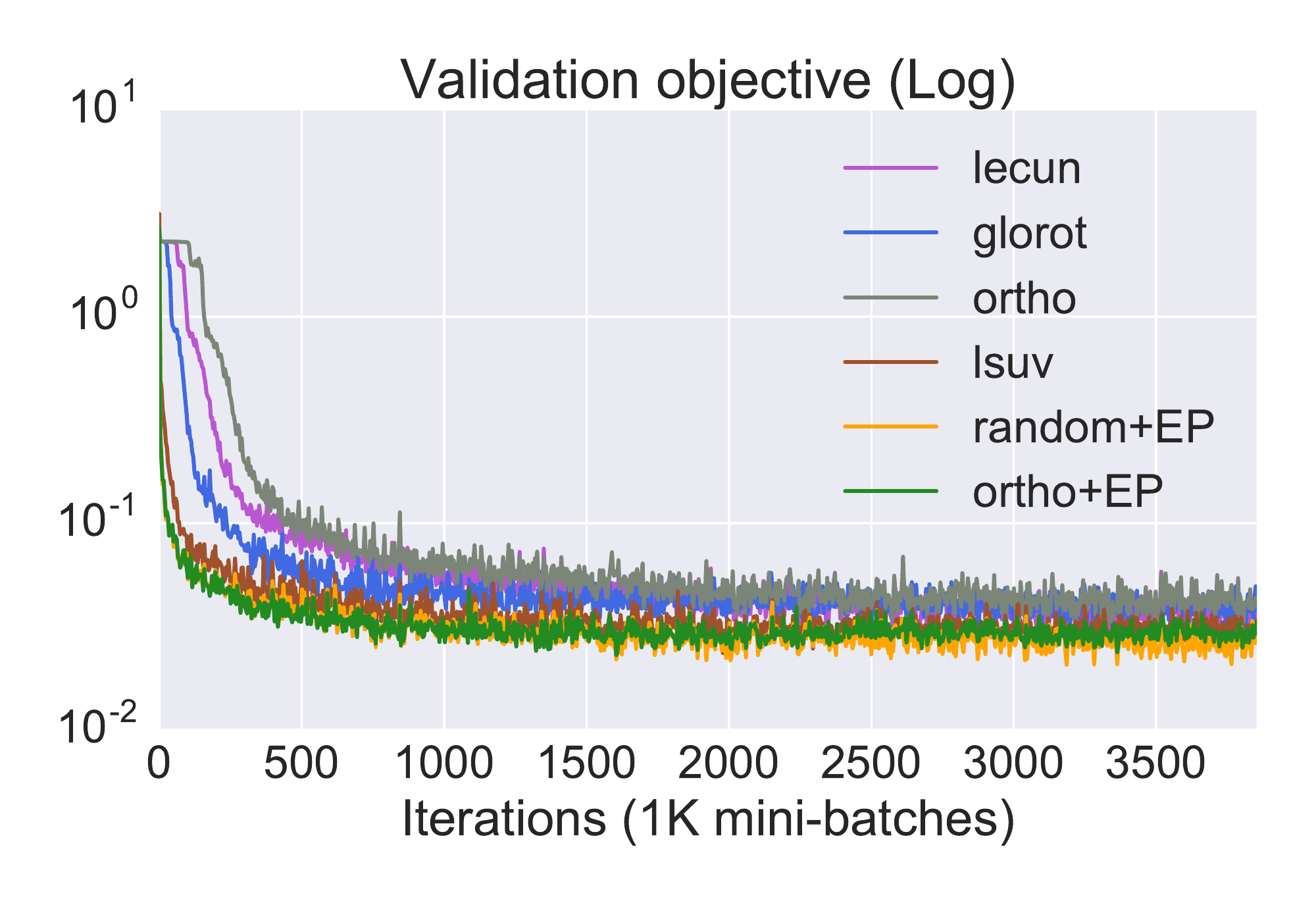}}
    \hfil
    \caption{Learning curves for training and validation objectives on MNIST (logarithmic scale): (a) and (d) results with shallow network, (b) and (e) results with deep network,
    (c) and (f) results with deep network with data augmentation.}
    \label{fig:mnist}
\end{figure*}
\begin{table}[!t]
\caption{Quantitative results on MNIST with shallow network~\cite{simard2003best} and different initialization strategies}
\label{tab:mnistshallow}
\centering
\begin{tabular}{p{1.4cm}cp{1.5cm}}
    \textbf{Method  (*sigmoid)} & \textbf{T.E. (\%)} & \textbf{Train Time ($10^3$ secs)}\\
    \hline
    lecun* & 1.91$\pm$0.03 & 2.0 \\
    glorot* & 1.86$\pm$0.04 & 2.0 \\
    ortho* & 1.86$\pm$0.02 & 2.0 \\
    lsuv* & 1.87$\pm$0.01 & 2.0 \\
    ortho+EP* & 1.89$\pm$0.02 & 2.0 \\
    random+EP* & \textbf{1.78$\pm$0.06} & 2.0 \\
    \hline
\end{tabular}
\end{table}
\begin{table}[!t]
\caption{Quantitative results on MNIST with shallow network~\cite{simard2003best} and dropout ($p=0.5$) for different initialization strategies}
\label{tab:mnistshallowdropout}
\centering
\begin{tabular}{p{1.4cm}cp{1.5cm}}
    \textbf{Method  (*sigmoid)} & \textbf{T.E. (\%)} & \textbf{Train Time ($10^3$ secs)} \\
    \hline
    \cite{simard2003best} & \textbf{1.60} & - \\
    lecun* & 1.65$\pm$0.03 & 6.0 \\
    glorot* & 1.63$\pm$0.02 & 6.0 \\
    ortho* & \textbf{1.62$\pm$0.04} & 6.0 \\
    lsuv* & \textbf{1.61$\pm$0.03} & 6.0 \\
    ortho+EP* & \textbf{1.61$\pm$0.03} & 6.0 \\
    random+EP* & \textbf{1.59$\pm$0.03} & 6.0 \\
    \hline
\end{tabular}
\end{table}
In this section, we compare different initialization methods for a shallow network with 800 hidden units 
characterized by sigmoid activation functions and with softmax output layer on the MNIST bechmark dataset~\cite{lecun1998gradient}. 
Mini-batch gradient descent with momentum 
equal to $0.9$ and learning rate equal to $0.001$ is applied to the minimization of the cross-entropy 
objective function. The size of each mini-batch consists of 50 training samples and the training algorithm
is run for $900k$ iterations. All results, including the learning curves, are averaged over 4 different random 
initializations. Data is normalized and mean-centered to lie in the range $[-1,1]$.

We plot the training curves for the objective computed on both the training and the validation sets, see Fig.~\ref{fig:mnist}. 
We see that our proposed initialization scheme allows to converge faster even in the case of non-deep models. 
Furthermore, our strategy is quite robust to the initial generation of weights. 
Previous attempts have shown that random initializations produces solutions with very different performance,
thus requiring very expensive pre-processing strategies to reduce the variance of generalization estimates, 
like unsupervised pre-training~\cite{erhan2010does}.
To the best of our knowledge, 
this is the very first time that random initialization without unsupervised pre-training allows to converge 
to solutions with reduced variance in performance (see \textit{random+EP} Fig.~\ref{fig:mnist}a and 
Fig.~\ref{fig:mnist}d). 

In Table~\ref{tab:mnistshallow}, we show the test errors with the related training times.\footnote{All training 
times are equal, since they represent the time to perform all training epochs.} 
Model selection is performed based on the minimization of the validation objective. 
The performance in terms of generalization are pretty similar for all methods, thus the main advantage of our strategy
consists of faster convergence.

We apply also dropout \cite{srivastava2014dropout} (where the dropping probability is set to $0.5$) and compare with
 the state of the art results reported in~\cite{simard2003best}, which are obtained 
with the same network architecture, but with hand-crafted activation functions. In particular, the authors
use squashed hyperbolic tangent activations, defined as $f(z)=A tanh(Bz)$, where $A=1.7159$ and $B=2/3$ 
(the parametrization derives by the requirement that $f(1)=1$ and $f(-1)=-1$), because their function gain 
is close to $1$ in the nominal region and the computed gradients are therefore less attenuated~\cite{lecun1998gradient}
compared to sigmoids. 
Table~\ref{tab:mnistshallowdropout} summarizes the results. To the best of our knowledge, this is the
first experimental trial demonstrating that sigmoids achieve similar performance to hyperbolic tangents~\cite{simard2003best}.

\subsection{MNIST: Deep Network}
\begin{table}[!t]
\caption{Quantitative results on MNIST with deep network~\cite{ciresan2010deep} and different initialization strategies}
\label{tab:mnistdeep}
\centering
\begin{tabular}{p{1.4cm}cp{1.5cm}}
    \textbf{Method  (*sigmoid)} & \textbf{T.E. (\%)} & \textbf{Train Time ($10^3$ secs)} \\
    \hline
    lecun* & 3.11$\pm$0.14 & 7.3 \\
    glorot* & 2.94$\pm$0.08 & 7.3 \\
    ortho* & 3.32$\pm$0.11 & 7.3 \\
    lsuv* & 2.07$\pm$0.07 & 7.3 \\
    ortho+EP* & \textbf{1.85$\pm$0.07} & 7.3 \\
    random+EP* & \textbf{1.92$\pm$0.08} & 7.3 \\
    \hline
\end{tabular}
\end{table}
\begin{figure*}[!t]
\centering
    \subfloat[]{\includegraphics[width=0.32\textwidth]{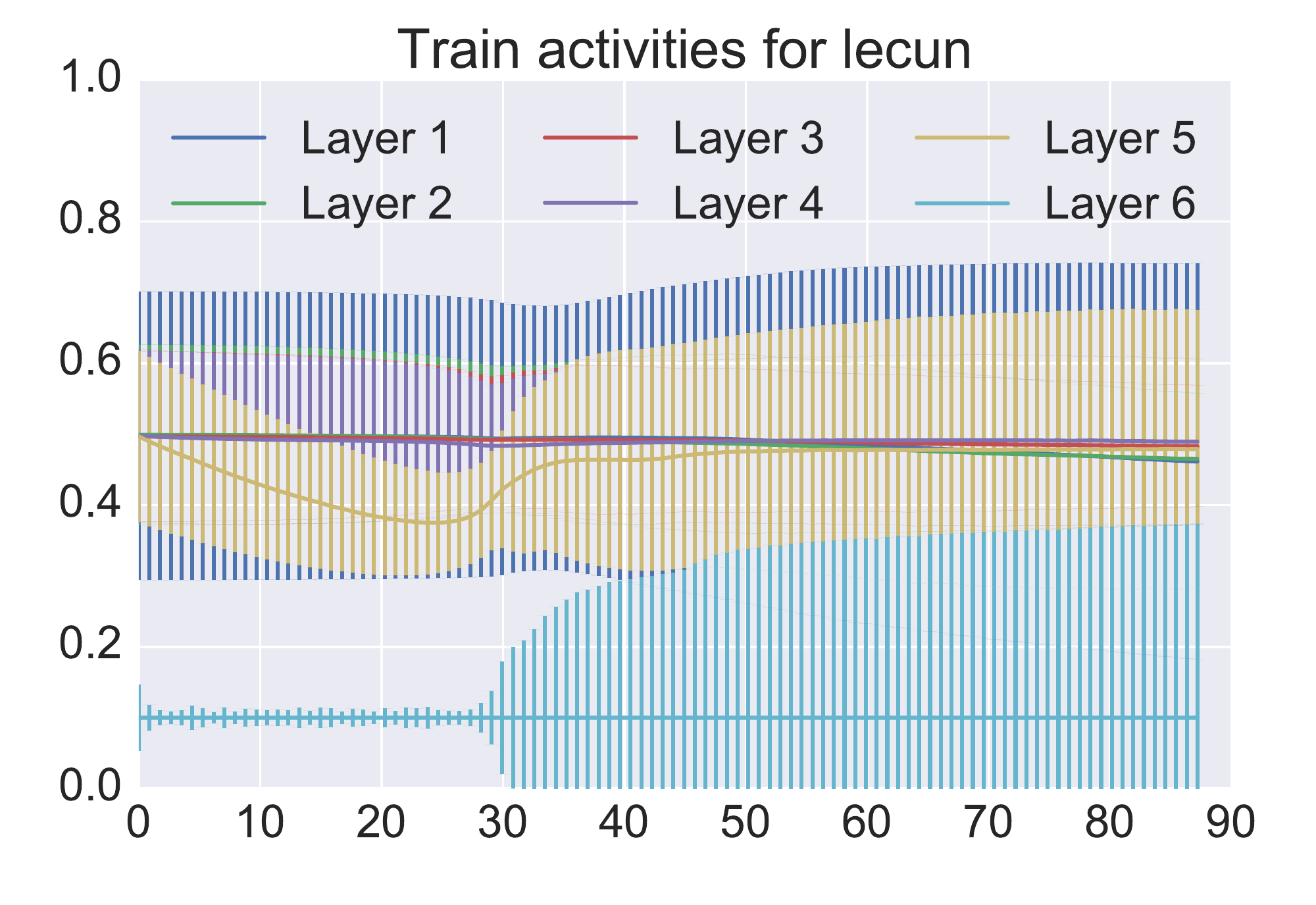}}
    \hfil
    \subfloat[]{\includegraphics[width=0.32\textwidth]{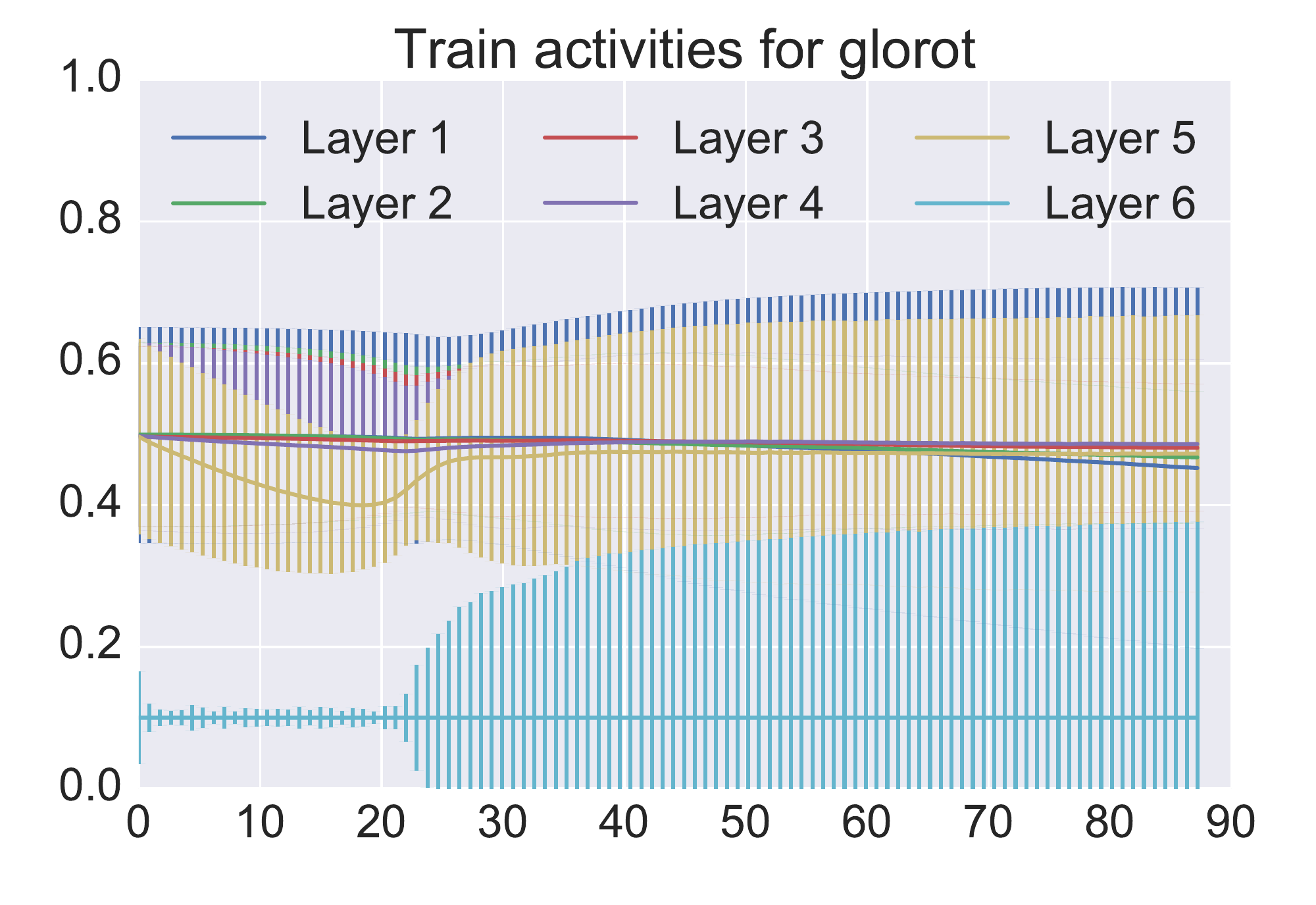}}
    \hfil
    \subfloat[]{\includegraphics[width=0.32\textwidth]{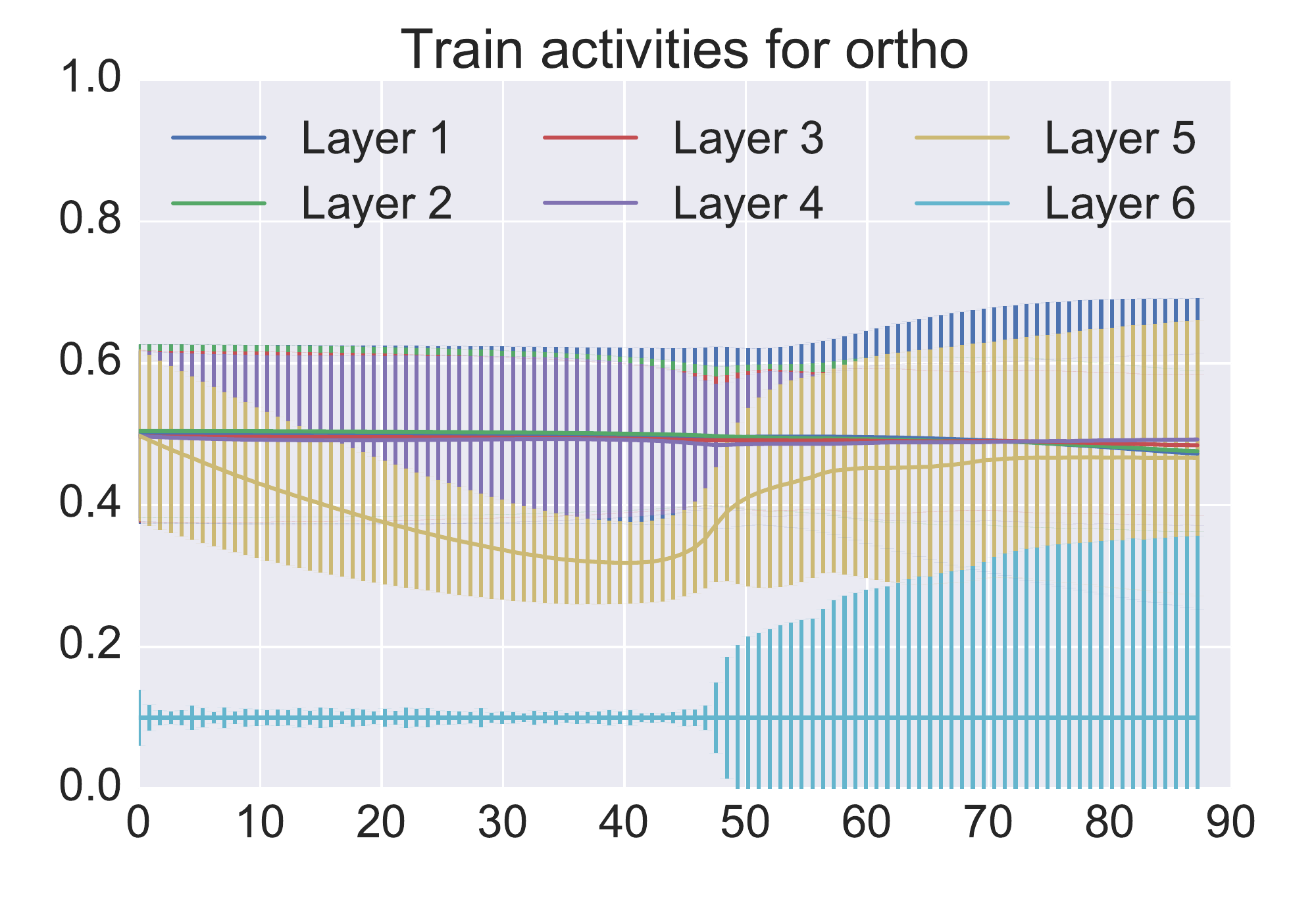}}
    \hfil
    \subfloat[]{\includegraphics[width=0.32\textwidth]{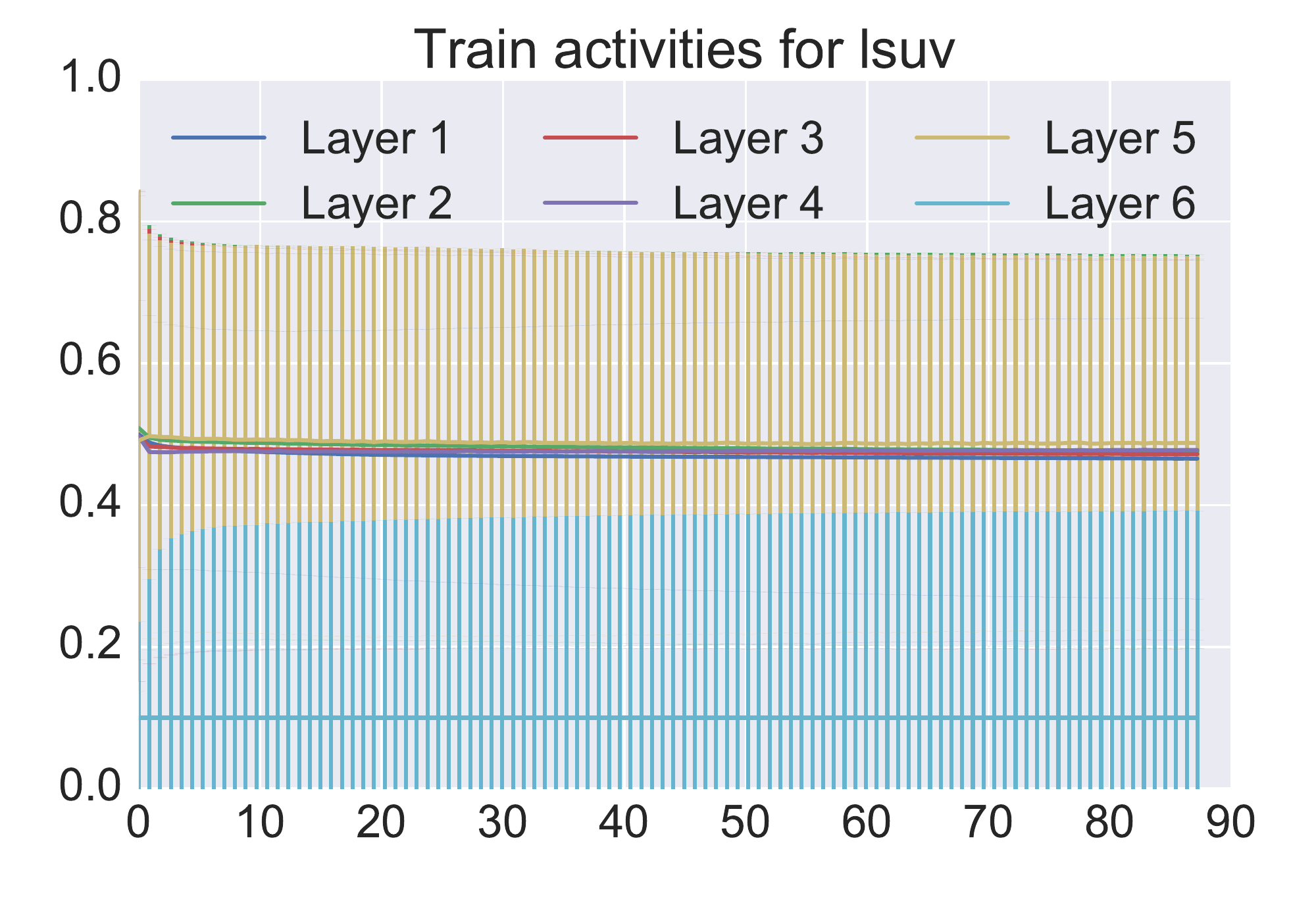}}
    \hfil
    \subfloat[]{\includegraphics[width=0.32\textwidth]{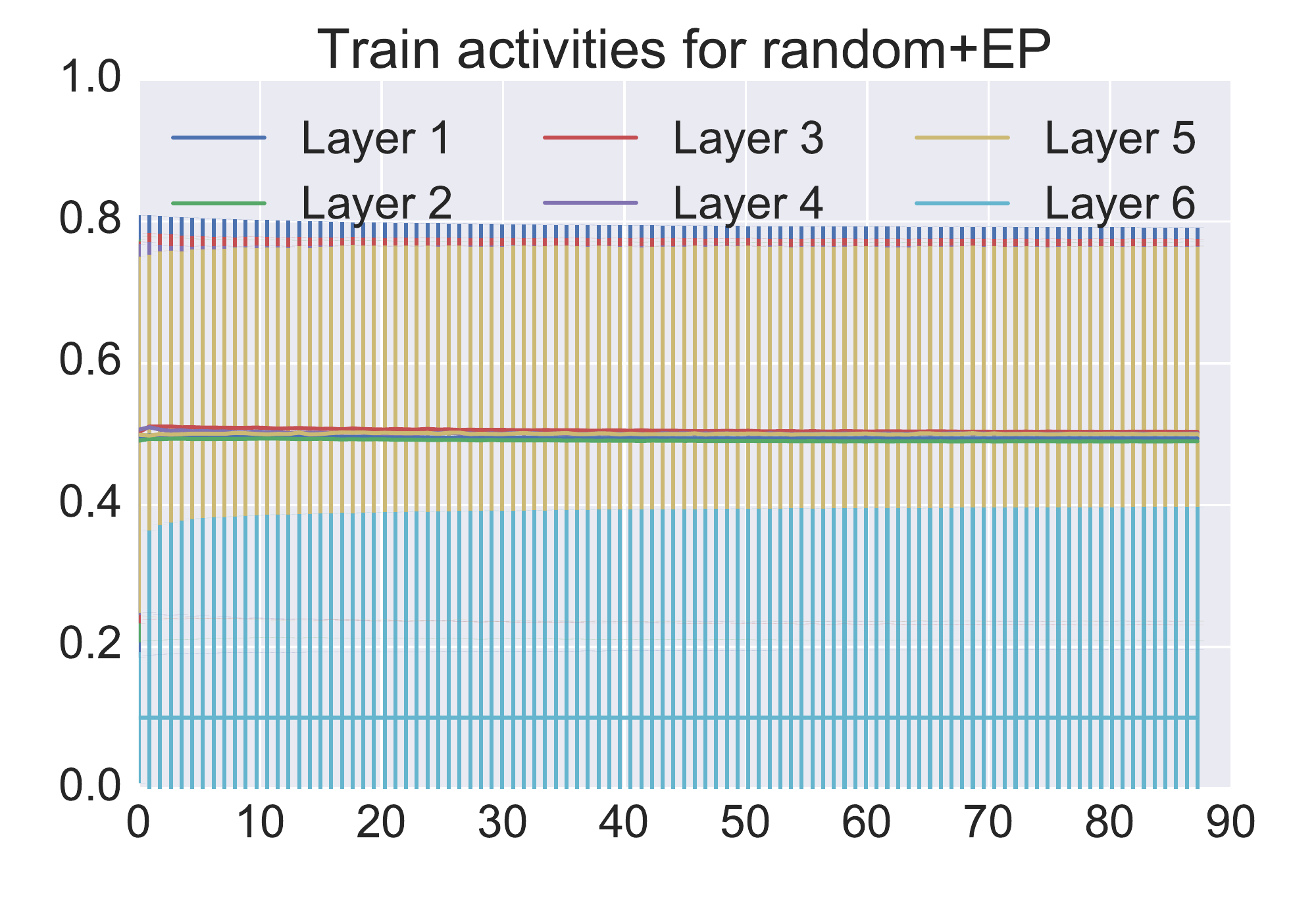}}
    \hfil
    \subfloat[]{\includegraphics[width=0.32\textwidth]{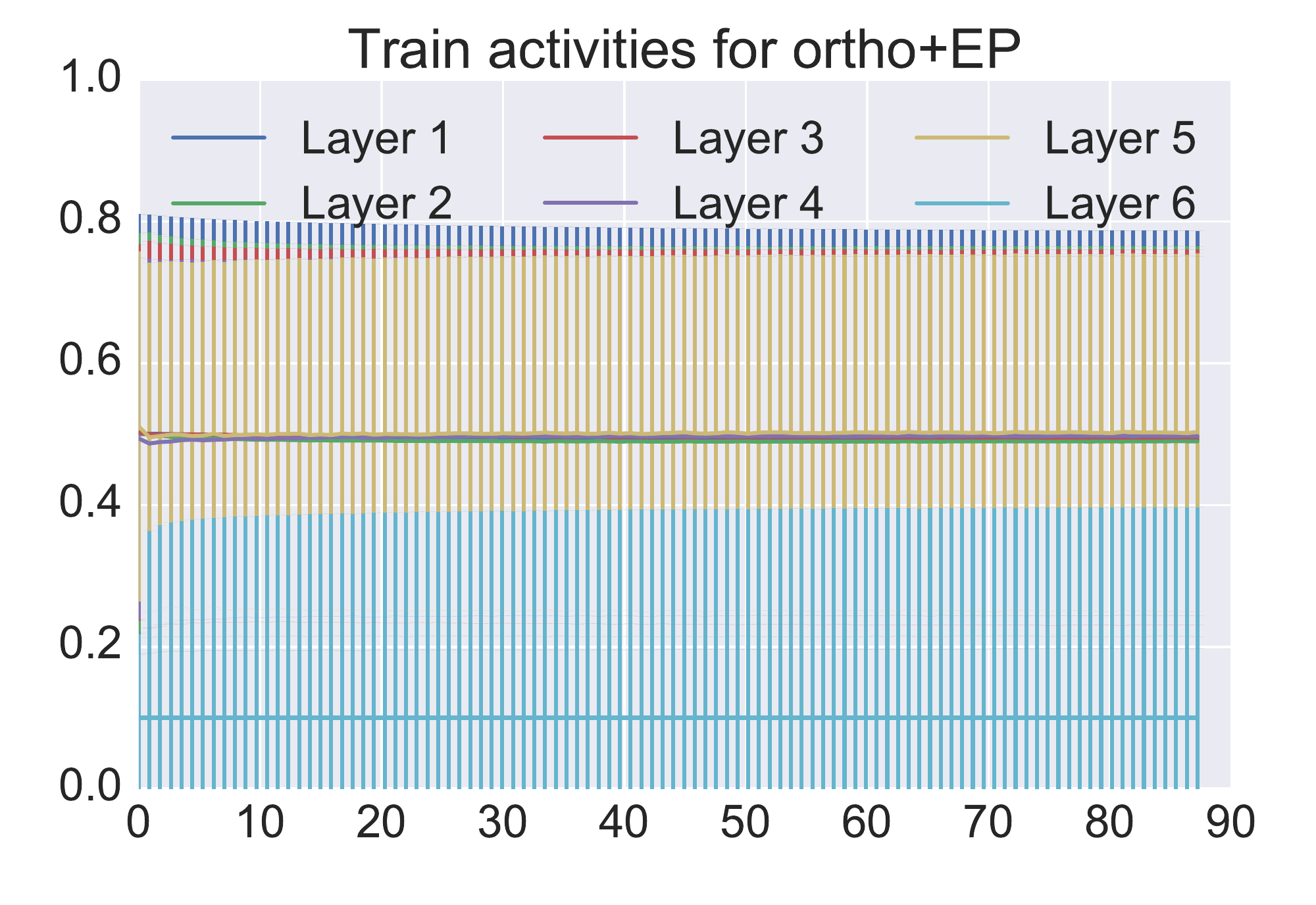}}
    \caption{Temporal evolution of statistics (mean and standard deviation) of activations in each layer for different initialization methods
    over the first $90k$ iterations.}
    \label{fig:vanishing}
\end{figure*}
\begin{figure*}[!t]
    \centering
        \subfloat[]{\includegraphics[width=0.32\textwidth]{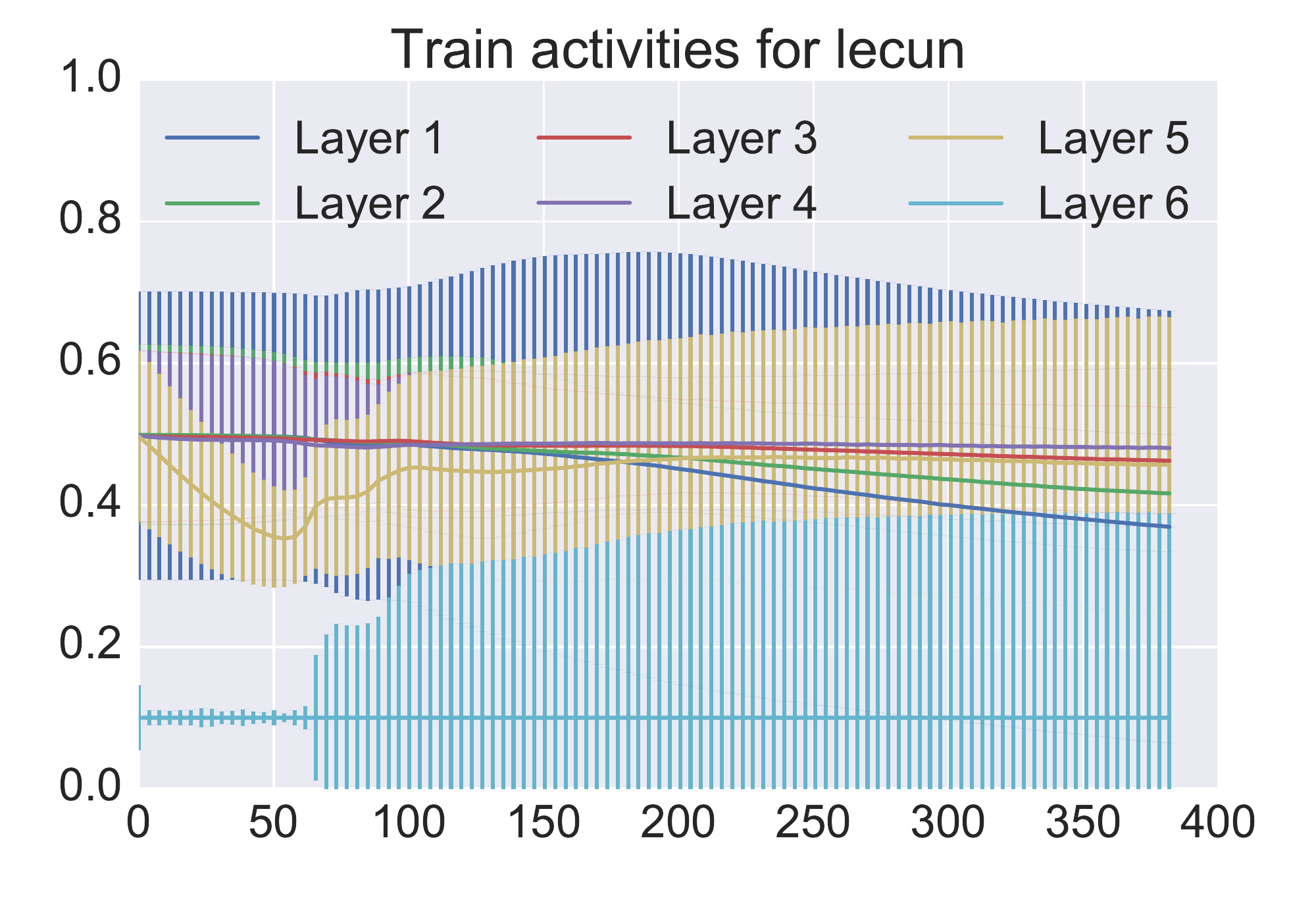}}
        \hfil
        \subfloat[]{\includegraphics[width=0.32\textwidth]{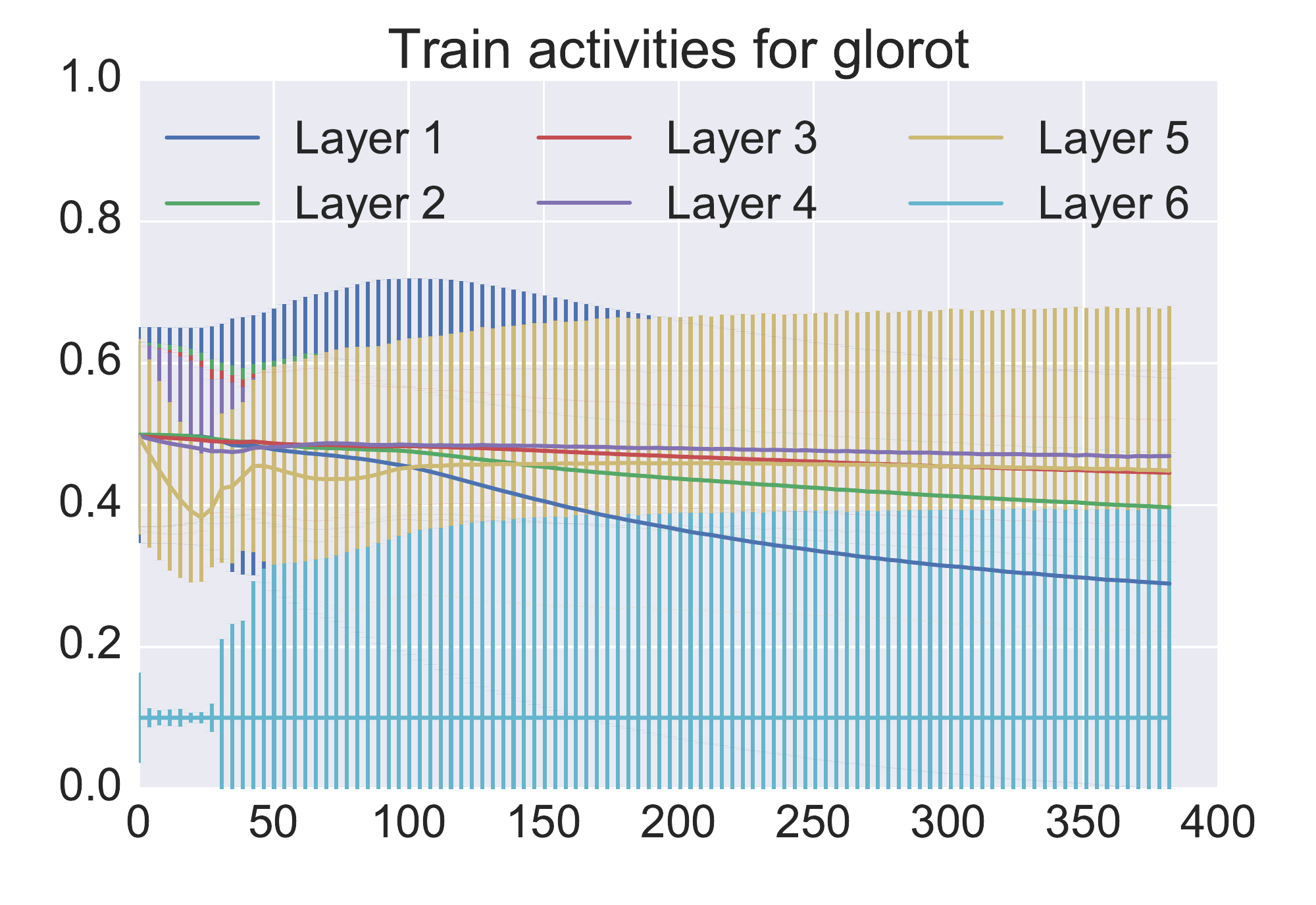}}
        \hfil
        \subfloat[]{\includegraphics[width=0.32\textwidth]{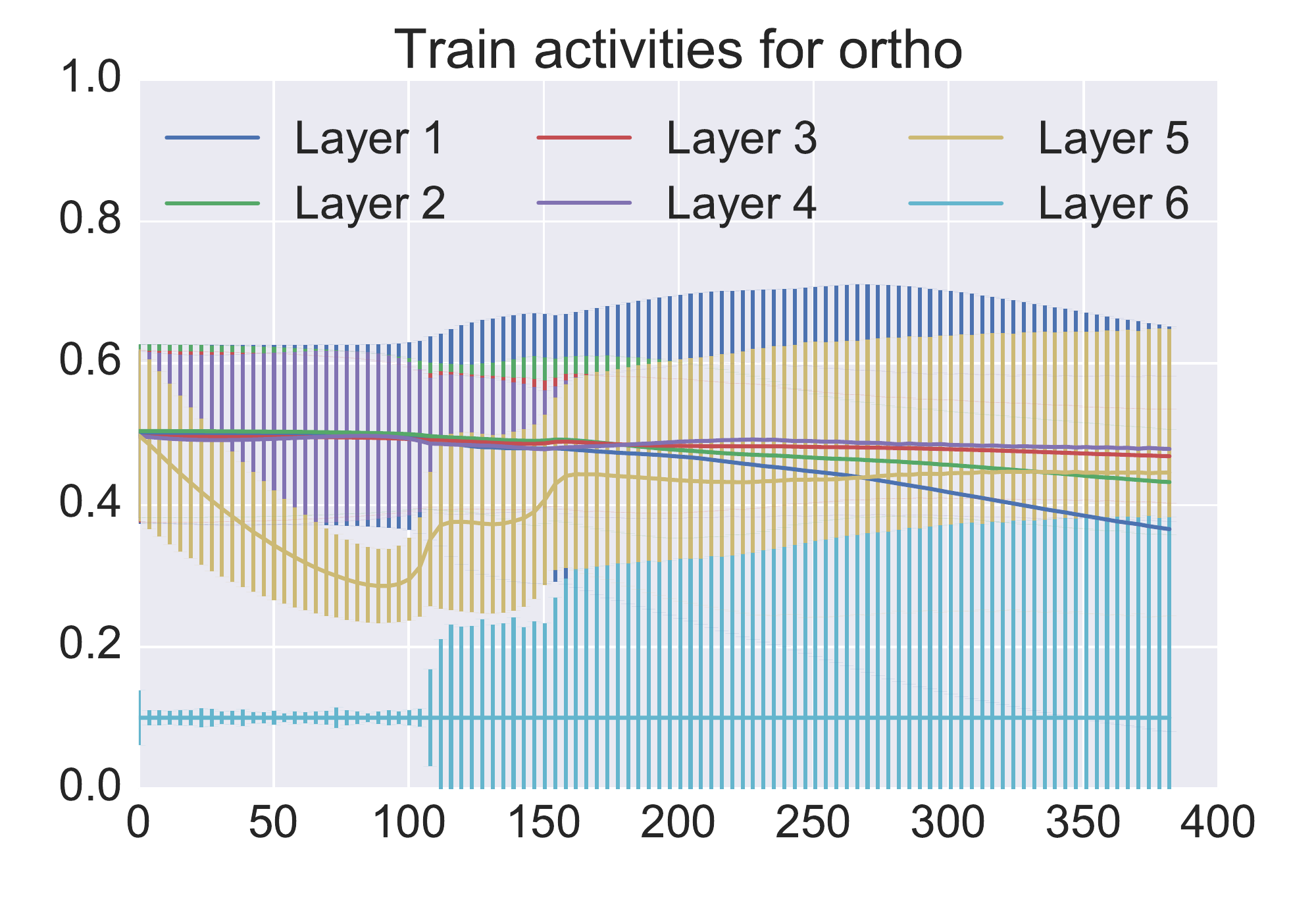}}
        \hfil
        \subfloat[]{\includegraphics[width=0.32\textwidth]{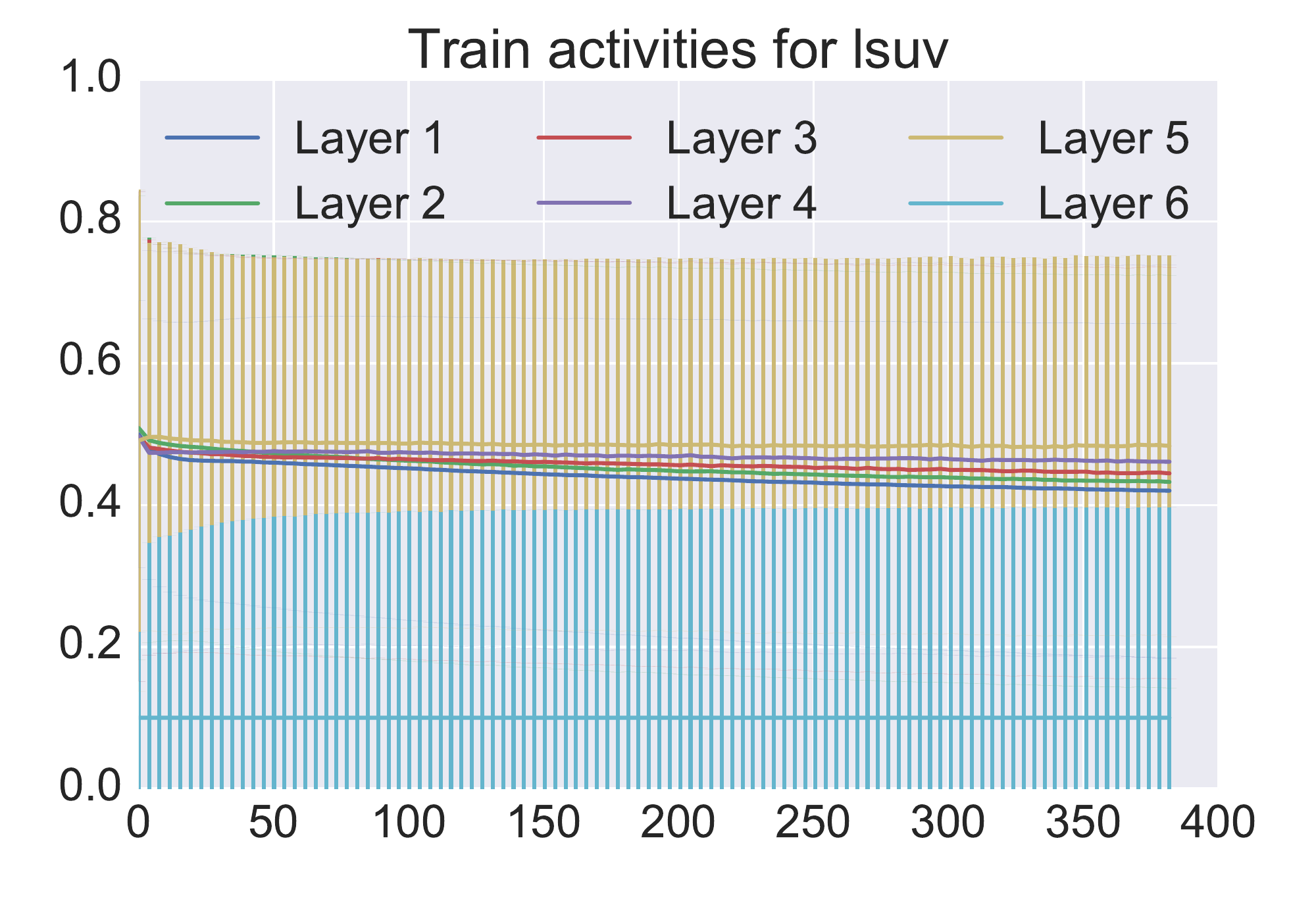}}
        \hfil
        \subfloat[]{\includegraphics[width=0.32\textwidth]{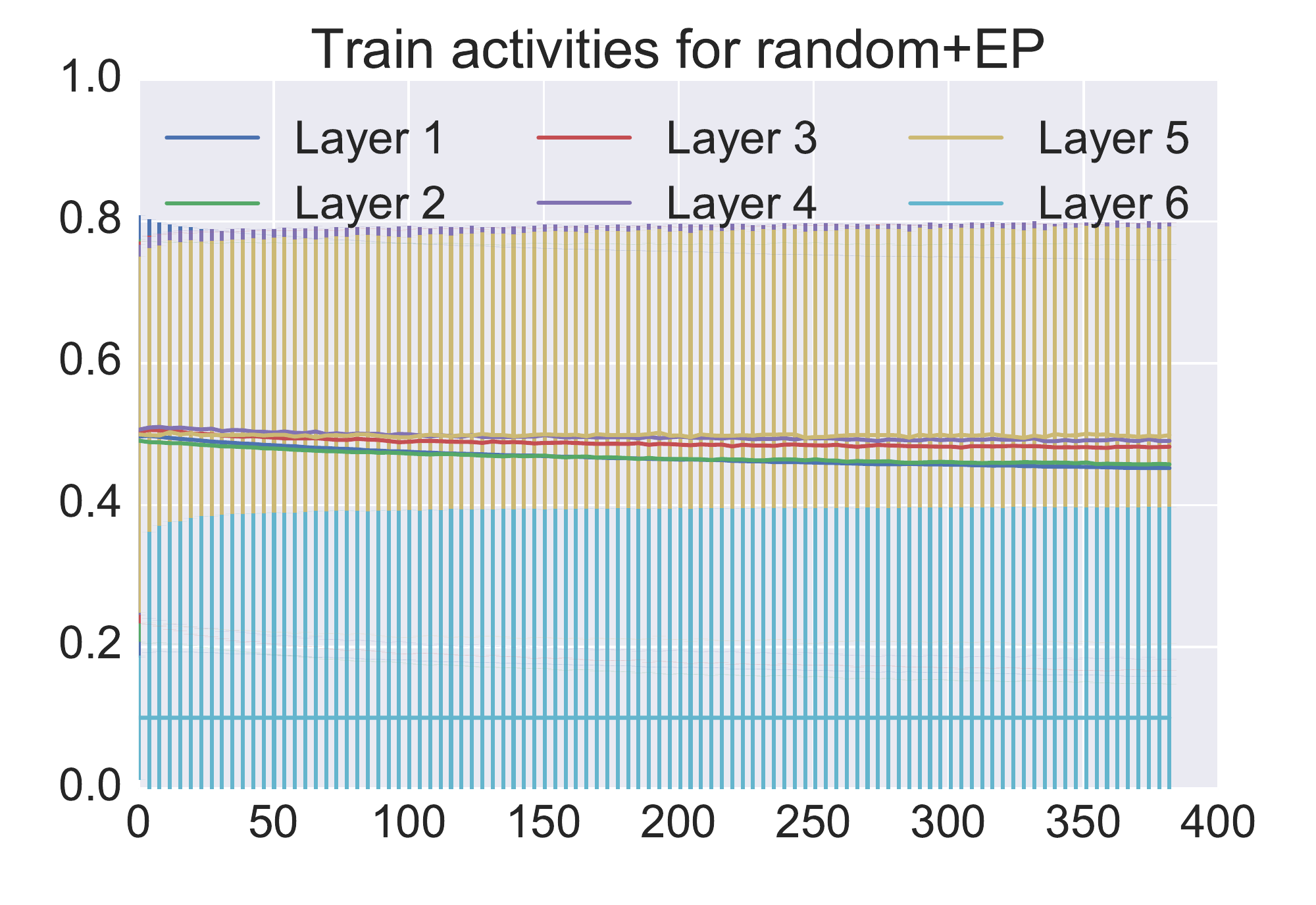}}
        \hfil
        \subfloat[]{\includegraphics[width=0.32\textwidth]{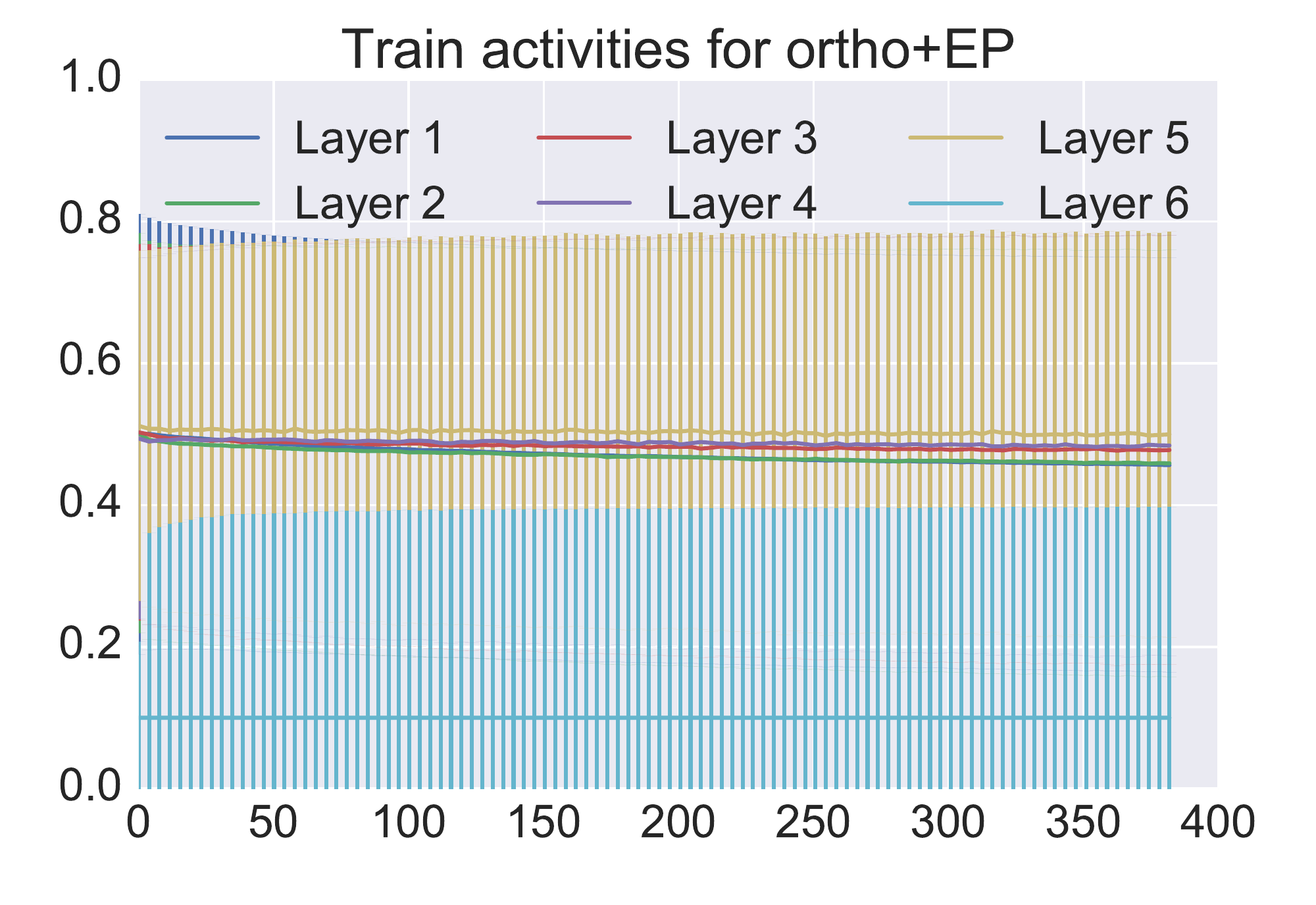}}
        \caption{Temporal evolution of statistics (mean and standard deviation) of activations in each layer for different initialization methods 
        over the first $90k$ iterations. Data augmentation is applied to improve performance.}
        \label{fig:vanishing2}
    \end{figure*}
We repeat the experiments of previous subsection with same setup, but with a deeper network. In particular, we use 
the same architecture of~\cite{ciresan2010deep}, consisting of 6 layers (5 hidden and 1 output layer), 
where the number of neurons in each layer is chosen according to a pyramidal structure, namely 2500, 2000, 
1500, 1000, 500 and 10 neurons, respectively.

From Fig.~\ref{fig:mnist}b and Fig.~\ref{fig:mnist}e, we observe immediately that our initialization provides 
a gain in terms of both convergence speed and the quality of the obtained solution. It is also interesting 
to see that the competitors observe very slow learning in the early stages of training, probably due to the fact 
that the network parameters are initialized on flat regions of the training objective and the gradients are
therefore vanishing. 

We also analyze the evolution of the statistics of each layer over the first $90k$ iterations (we consider
the mean and the standard deviation of each layer obtained by averaging neuron activities over mini-batches). 
Fig.~\ref{fig:vanishing} shows the behaviour of the network for different initialization strategies.
For almost all competitors, the activities in the last hidden layer (layer 5) tend to be biased towards zero
and no significant variation is appreciated in the output layer. This is a symptomatic behaviour, that have 
been already observed in~\cite{glorot2010understanding} for networks with standard logistic activations. 
It is interesting to mention that this problem is not visible for \textit{lsuv} and our proposed initializations.
This is probably due to the fact that both impose some conditions on the variance of the weights of each neuron.
In particular, \textit{lsuv} imposes a unit variance, while our conditions require a larger value, which is
based on an information-theoretically criterion. This has also some beneficial impact on the generalization
performance as shown in Table~\ref{tab:mnistdeep}. Therefore, our weight initialization 
is sufficient to guarantee a better propagation of gradients at all layers for the whole training process, 
allowing to achieve faster convergence and better solutions.

\begin{table}[!t]
\caption{Quantitative results on MNIST with deep network~\cite{ciresan2010deep} and data augmentation for different initialization strategies}
\label{tab:mnistaugmentation}
\centering
\begin{tabular}{p{1.4cm}cp{1.5cm}}
     \textbf{Method  (*sigmoid)} & \textbf{T.E. (\%)} & \textbf{Train Time ($10^3$ secs)} \\
     \hline
     \cite{ciresan2010deep} & 0.32 & 412.2 \\
     \cite{ciresan2010deep}$^+$ & \textbf{0.62} & 35.4 \\
     lecun* & 0.80 & 35.4 \\
     glorot* & 0.82 & 35.4 \\
     ortho* & 0.80 & 35.4 \\
     lsuv* & 0.79 & 35.4 \\
     ortho+EP* & 0.81 & 35.4 \\
     random+EP* & \textbf{0.68} & 35.4 \\
     \hline
     \multicolumn{3}{l}{$^+$ Our implementation}
\end{tabular}
\end{table}

We apply also data augmentation to compare against the state of the art results reported in~\cite{ciresan2010deep} with squashed hyperbolic 
tangent activations. In particular, training data are augmented following the methodology suggested in~\cite{ciresan2010deep}.
We use elastic distortion, to emulate uncontrolled oscillations of hand muscles ($\sigma=5.5$ and $\alpha=37$,
see~\cite{ciresan2010deep}),\footnote{Here, $\sigma$ does not refer to $\sigma$ introduced in Section~\ref{sec:statistical}.} 
rotation (with an angle randonly sampled from $[-\beta,\beta]$, where $\beta=7.5°$ 
for digits 1 and 7 and $\beta=15°$ for all other digits) and horizontal and vertical scaling (with scaling factor 
randomly sampled from $[1-\gamma/100,1+\gamma/100]$, where $\gamma=17.5$). We also rerun the experiments of~\cite{ciresan2010deep}
and use these as baseline, since we were not able to replicate the results.

The learning curves in Fig.~\ref{fig:mnist}c and Fig.~\ref{fig:mnist}f confirms the findings that our method achieves faster
convergence and better solutions. It is important to mention that the use of data augmentation plays an important role in
achieving better generalization performance and that it can partially overcome to problems incurred by using a wrong initialization. 
Nevertheless, the advantages of our initialization are still visible.
Fig.~\ref{fig:vanishing2} shows the temporal evolution of statistics for each layer, emphasizing the fact that almost
all competitors are subject to the problem of saturation of the last hidden layer in the early stages of training~\cite{glorot2010understanding}.
Table~\ref{tab:mnistaugmentation} summarizes the quantitative results. Our initialization strategy (\textit{random+EP})
allows to achieve 0.68\% of test error rate, which is very close to the performance (namely 0.62\%) obtained by using the hand-crafted 
hyperbolic tangent of~\cite{ciresan2010deep}. This provides further evidence that deep networks with standard logistic activations can
perform similarly to networks with hyperbolic tangents and that the training is made feasible through some simple conditions 
at initialization.

\subsection{Copying Memory Problem: Recurrent Neural Network}
\begin{figure}[!t]
\centering
    \subfloat[]{\includegraphics[width=0.7\linewidth]{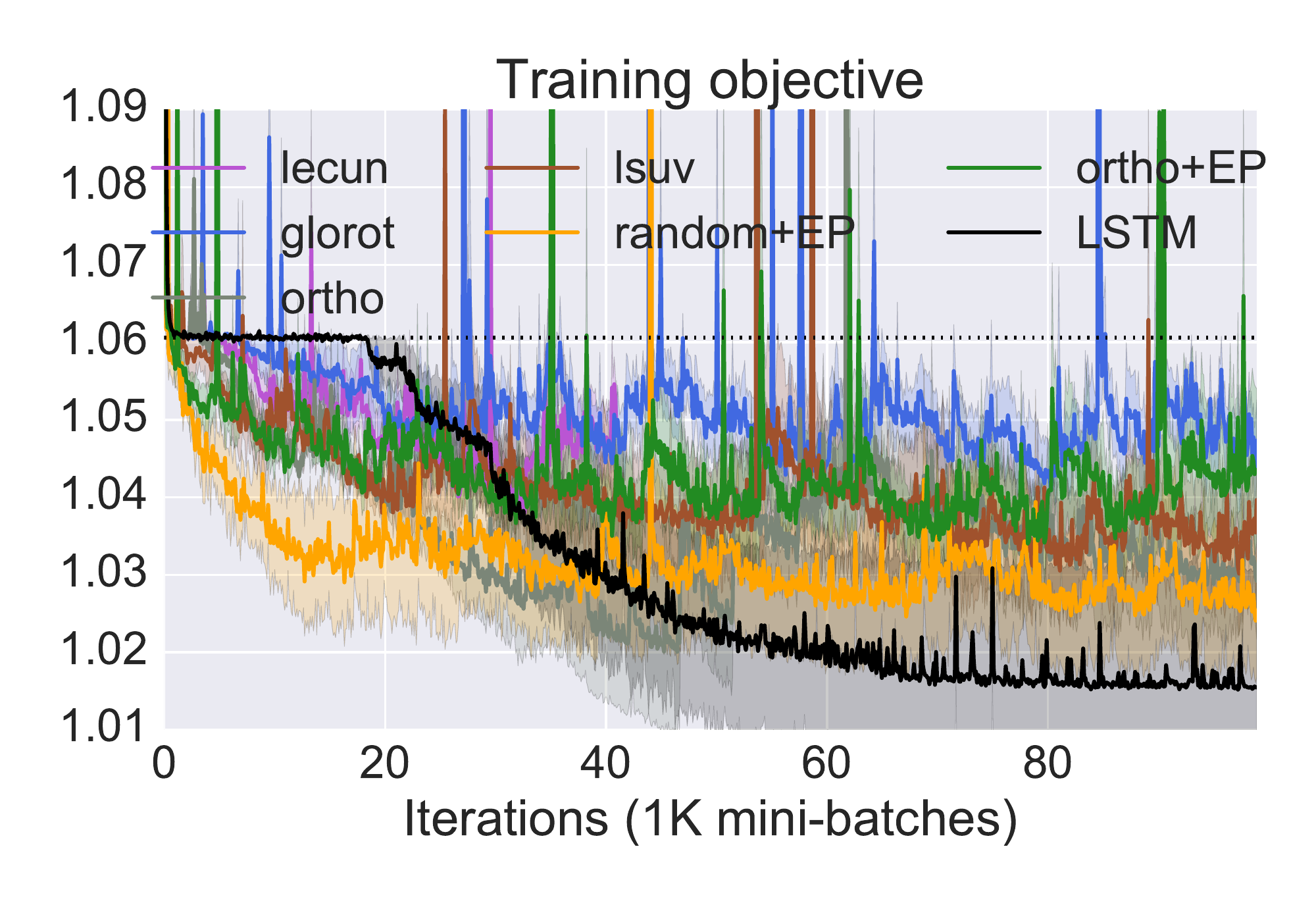}}
    \hfil
    \subfloat[]{\includegraphics[width=0.7\linewidth]{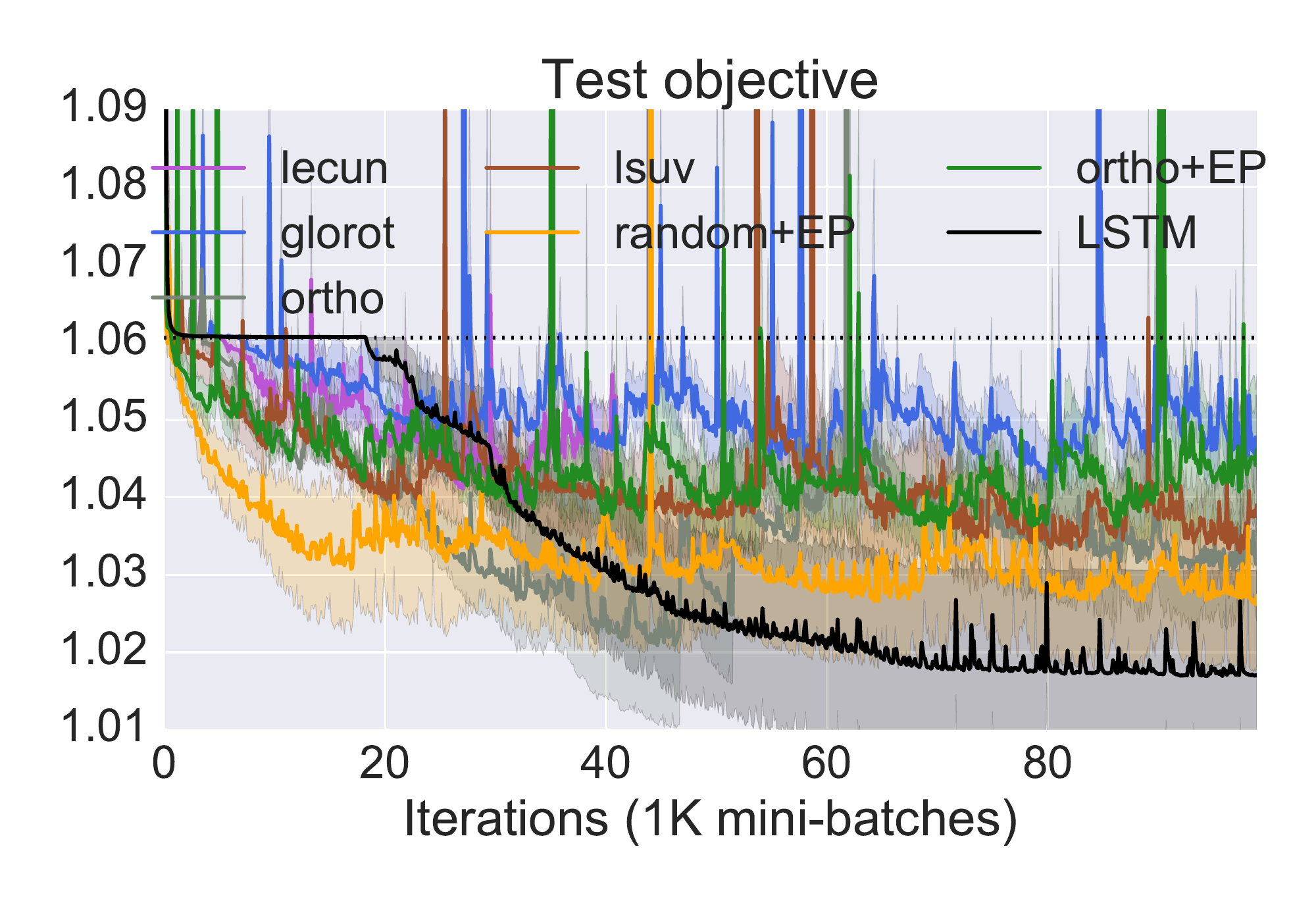}}
    \caption{Training and test learning curves on the copy memory problem for different strategies.}
    \label{fig:copy}
\end{figure}
\begin{figure}[!t]
    \centering
        \includegraphics[clip,trim=5cm 12cm 5cm 12cm, width=\linewidth]{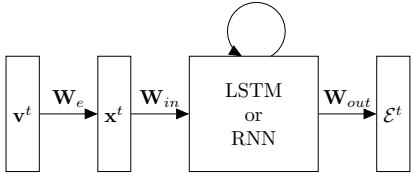}
        \caption{General architecture used in the experiments with recurrent neural networks. 
        $\mathbf{v}^t$, $\mathbf{\mathcal{E}}^t$ are the input and the output vectors, respectively,
        representing a charater or a word using one-hot encoding, while $\mathbf{x}^t$ represents 
        the distributed embedding of the input vector.} 
        \label{fig:architecture}
\end{figure}
In this section, we go even deeper and use RNNs as case study to compare the different initialization
strategies. It is well known that these models have difficulties to remember information about inputs for 
long time intervals. This is due to the fact that during training, RNNs are affected by the problem of 
vanishing/exploding gradients~\cite{bengio1994learning,pascanu2013difficulty}.
The literature contains a plethora of proposed solutions. In particular, the authors 
in~\cite{jaeger2004harnessing,lukovsevivcius2009reservoir} propose a strategy called Echo-State Networks (ESN), 
which consist in carefully initializing the recurrent weights and training only the output parameters. In practice, 
the recurrent weight matrix is initialized to have a spectral radius close to one, such that inputs can "echo" for 
long time. This represents a very drastic solution, which doesn't exploit the full potential of RNNs.
Authors in~\cite{martens2011learning} show that long-term dependencies can be learnt using Hessian-free
optimization, which is provided with information about the curvature of the loss surface and is therefore
able to deal with vanishing gradients. Authors
in~\cite{pascanu2013difficulty} propose a solution to train RNNs, where a regularizer, specifically designed to cope with
vanishing gradients, is applied to the loss objective and a simple momentum-based optimizer is used
in combination with gradient clipping, that ensures that gradients do no explode. In our experiments, we use the
same strategies of~\cite{pascanu2013difficulty} and we focus on analyzing the impact of initialization on this
problem. It is also important to mention that there is a recent line of research in RNNs, studying unitary recurrent
weight matrices, see for example~\cite{arjovsky2016unitary,jing2017tunable,mhammedi2017efficient}. Nevertheless, in this work
we want to study the most general case, where the feasible set consists of the whole parameter space.

In order to study the capability of models in learning long-term dependencies, we consider a similar pathological 
task introduced for the first time in~\cite{hochreiter1997long}, called the copy memory problem. In this task, we 
are given a dictionary of ten characters, viz. $\{a_i\}_{i=0}^9$. The input is a $T+6$ length sequence
containing characters from the dictionary. Specifically, the first three characters in the sequence are drawn 
uniformly and independently with replacement from the subset $\{a_i\}_{i=0}^7$ and represent the sequence 
to be memorized. The next $T-1$ characters are set to $a_8$ and represent a dummy sequence. The next character
is set to $a_9$ and represent a trigger to inform the model that it should start to predict the memorized 
sequence. The last three characters are set to $a_8$ and represent a dummy sequence. The ground truth output
is a $T+6$ length sequence, where the first $T+3$ entries are set to $a_8$ and the last three characters are
the copy of the first characters in the input sequence. Therefore, the task consists in memorizing the input
sequence for $T$ time instants and then output the memorized sequence. In the experiments, we set $T=100$ and 
generate training and test datasets consisting of $1000$ samples each.

We compare RNNs using different initialization strategies and report also the results of two baselines. 
The first baseline consists in outputting $a_8$ for the first $T+3$ entries and randomly sampling from the 
subset $\{a_i\}_{i=0}^7$ for the last three characters. This is equivalent to having a \textit{memoryless} 
strategy. The second baseline consists in predicting the output using a LSTM model~\cite{hochreiter1997long},
which is the most widely used alternative to RNNs and is able to learn long-term dependencies. Performance
are measured in terms of perplexity and this quantity is also used as training objective. 
Note that the perplexity of the first baseline can be analytically
computed and is equal to $\exp{\big\{\frac{3\ln 8}{T+6}\big\}}\approx 1.06$.

The network architecture used in the experiments is shown in Fig.~\ref{fig:architecture}, where the input and 
the output vectors represent the one-hot encoding of any character in the dictionary.
The size of hidden layers in the recurrent model (which is also used to determine the size of the
embedding vector) depends whether we are using RNN or LSTM. In particular, RNN and LSTM have hidden layers of 
size $100$ and $52$, respectively, which is equivalent to have roughly $32000$ parameters per model.
Adam optimizer with learning rate equal to $0.001$ is used in training and the algorithm is run for $100k$ iterations. 
All results are averaged over 4 different random initializations. 

Figure~\ref{fig:copy} shows the learning curves for the different initialization strategies
as well as the learning curves of the two baselines. It is possible to see that eventually all approaches perform
better than the \textit{memoryless} strategy, meaning that the networks effectively learn to memorize some
information. Furthemore, our initialization (\textit{random+EP}) outperforms all other approaches during the
first $30k$ iterations, including LSTM. There is a phase between $30k$ and $50k$ iterations, in which \textit{ortho}
achieves the best performance. As discussed in~\cite{saxe2013exact}, an orthogonal initialization can guarantee
that the recurrent weight matrix remains orthogonal during the whole training process even in the case of nonlinear 
activation functions. This is particularly useful in the copying memory problem, since the family of orthogonal
matrices contain the solution, that allows the RNN to learn the identity function, namely to copy the exact 
input sequence to the output. Note that after $50k$ iterations the performance of \textit{ortho} degrades due to the problem of exploding gradients. It is interesting to note that at convergence we achieve the best performance among the other initializations and are able to get closer to the 
results of LSTM.
  
\subsection{Language modelling: Recurrent Neural Network}
\begin{table}[!t]
\caption{Quantitative results on the language modelling task for different strategies.}
\label{tab:lm}
\centering
\begin{tabular}{p{1.4cm}cp{1.5cm}}
    \textbf{Method  (*sigmoid)} & \textbf{T.P.} & \textbf{Train Time ($10^3$ secs)} \\
    \hline
    lstm & \textbf{128.3$\pm$0.5} & 77.0 \\
    lecun* & 143.1$\pm$0.8 & 28.7 \\
    glorot* & 155.1$\pm$0.9 & 28.7 \\
    ortho* & 156.1$\pm$0.5 & 28.7 \\
    lsuv* & \textbf{135.9$\pm$0.1} & 28.7 \\
    ortho+EP* & \textbf{135.5}$\pm$0.3 & 28.7 \\
    random+EP* & 164.0$\pm$0.5 & 28.7 \\
    \hline
\end{tabular}
\end{table}
\begin{table}[!t]
\caption{Quantitative results on the language modelling task for different strategies with dropout.}
\label{tab:lm_dropout}
\centering
\begin{tabular}{p{1.4cm}cp{1.5cm}}
    \textbf{Method  (*sigmoid)} & \textbf{T.P.} & \textbf{Train Time ($10^3$ secs)} \\
    \hline
    lstm & \textbf{128.7$\pm$0.4} & 80.0 \\
    lecun* & 147.8$\pm$0.6 & 32.3 \\
    glorot* & 159.2$\pm$0.4 & 32.3 \\
    ortho* & 161.2$\pm$0.3 & 32.3 \\
    lsuv* & 139.8$\pm$0.4 & 32.3 \\
    ortho+EP* & 139.3$\pm$0.3 & 32.3 \\
    random+EP* & \textbf{130.1$\pm$0.4} & 32.3 \\
    \hline
\end{tabular}
\end{table}
In this section, we conduct experiments on a real-world task, namely the language
modelling problem, using the Penn Tree Bank (PTB) dataset~\cite{marcus1993building}.
The dataset consists of $929k$ training words, $73k$ validation words, $82k$ test words
and the vocabulary has $10k$ words. The dataset is downloaded from Tomas Mikolov's
webpage.\footnote{http://www.fit.vutbr.cz/~imikolov/rnnlm/simple-examples.tgz}

We compare RNNs using different initialization strategies and report also the results of the
baseline using LSTM~\cite{hochreiter1997long}.
The aim of these experiments is to show that our conditions allow to improve the
generalization performance of RNNs. With our initialization, we are able to achieve comparable
performance to the one obtained by LSTM.

The network architecture is the same as the one used in the copying memory problem with
different size of the hidden layers. In particular, we use $200$ neurons for RNNs and 
$190$ neurons for LSTM. This corresponds to have roughly $4090k$ parameters for each model.

Adam optimizer with learning rate equal to $0.0001$ and gradient clipping~\cite{pascanu2013difficulty} 
is used for training the models and the algorithm is run for $500k$ iterations.

Results are averaged over 4 different random initializations and are shown in Table~\ref{tab:lm}
(the acronym T.P. stands for test perplexity). 
\textit{lsuv} and \textit{ortho+EP} achieve performance closer
to the ones of LSTM, while \textit{random+EP} obtains apparently very bad performance.
We argue that the generation of the initial matrices in \textit{random+EP} allows to
sample from a much larger space of matrices and consequently Adam has more chances to overfit. 
Note that the problem of overfitting of the Adam optimizer is known and is
discussed from a theoretical perspective in~\cite{wilson2017marginal}. To validate
this claim and potentially reduce the problem of overfitting, we run also the experiments with 
dropout following the procedure in~\cite{zaremba2014recurrent}, where the dropping probability
is set to $0.5$. Results, shown in Table~\ref{tab:lm_dropout}, confirm our initial claim. 
\textit{random+EP} outperforms all other initialization and is able to achieve performance 
comparable to LSTM. Note that RNNs offer a clear advantage in terms of computational complexity
over LSTM, as demonstrated by the training times in Table~\ref{tab:lm_dropout}.

%
{\section{Conclusion}\label{sec:conclusion}}
This work shows that a careful initialization of the parameters
is sufficient to successfully train feedforward neural networks with
standard logistic activations. The initialization is based on
some conditions, that are derived by studying the properties of a
single neuron through information theory. The study is corroborated
by numerous experiments over different known benchmarks and
different networks.

%
\appendices

{\section{Derivation of Bounds~(\ref{eq:bounds})}\label{sec:details}}
Recall that
\begin{equation}
  \label{eq:firstentropy}
  H(y) = H(z)+E_z\{\ln g'(z)\}
\end{equation}
where (we drop the extrema of integration for the sake of brevity in the notation)
\begin{IEEEeqnarray}{ll}
  H(z) &\doteq -\int\mathcal{N}(\mu,\sigma^2)\ln\mathcal{N}(\mu,\sigma^2)dz \nonumber\\
       &= \frac{1}{2\sigma^2}\int (z-\mu)^2\mathcal{N}(\mu,\sigma^2)dz + \ln(\sqrt{2\pi\sigma^2}) \nonumber\\
       &= \frac{1}{2} + \frac{1}{2}\ln(2\pi\sigma^2)
\end{IEEEeqnarray}
and
\begin{IEEEeqnarray}{ll}
  \label{eq:derivativeterm}
  E_z\{\ln g'(z)\} &\doteq \int\mathcal{N}(\mu,\sigma^2)\ln g'(z) dz \nonumber\\
                   &= \int \mathcal{N}(\mu,\sigma^2)\ln\big[g(z)\big(1-g(z)\big)\big] dz \nonumber\\
                   &= \int \mathcal{N}(\mu,\sigma^2)\ln\bigg[\frac{e^z}{(1+e^z)^2}\bigg] dz \nonumber\\
                   &= \int z\mathcal{N}(\mu,\sigma^2)dz -{2}\int \mathcal{N}(\mu,\sigma^2)\ln(1+e^z)dz \nonumber\\
                   &= \mu -{2}\int \mathcal{N}(\mu,\sigma^2)\ln(1+e^z)dz
\end{IEEEeqnarray}
Note that the integrand in~(\ref{eq:derivativeterm}) is always positive and that
$\max(0,z)<\ln(1+e^z)\leq \max(0,z)+\ln 2$ is true for any $z$. Therefore, we can easily
derive the following relations, namely:
\begin{equation}
   \label{eq:inequalities}
   A(\mu,\sigma)-2\ln 2 \leq E_z\{\ln g'(z)\} < A(\mu,\sigma)
\end{equation}
where
\begin{IEEEeqnarray}{ll}
    A(\mu,\sigma) &\doteq \mu - {2}\int_{-\infty}^{\infty}\mathcal{N}(\mu,\sigma^2)\max(0,z)dz \nonumber\\
                  &= \mu - {2}\int_{0}^{\infty}z\mathcal{N}(\mu,\sigma^2)dz \nonumber \\
                  &= \mu -\mu-\mu\:erf\bigg(\frac{\mu}{\sigma\sqrt{2}}\bigg)
                  - \frac{2\sigma}{\sqrt{2\pi}}e^{-\frac{\mu^2}{2\sigma^2}} \nonumber \\
                  &= -\mu\:erf\bigg(\frac{\mu}{\sigma\sqrt{2}}\bigg)
                  - \frac{2\sigma}{\sqrt{2\pi}}e^{-\frac{\mu^2}{2\sigma^2}}
\end{IEEEeqnarray}
By adding $H(z)$ to~(\ref{eq:inequalities}) and using~(\ref{eq:firstentropy}), we obtain that
\begin{IEEEeqnarray}{c}
  H(z)+A(\mu,\sigma)-2\ln 2 \leq H(y) < H(z)+A(\mu,\sigma) \nonumber\\
  H_B(\mu,\sigma) -2\ln 2 \leq H(y) < H_B(\mu,\sigma)
\end{IEEEeqnarray}
where $H_B(\mu,\sigma)\doteq H(z)+A(\mu,\sigma)$.

{\section{Proof of Theorem~\ref{th:stationarity}}\label{sec:proof}}
Recall the definition of the lower bound in~(\ref{eq:lowerbound}), namely:
\begin{equation}
  H_{B}(\mu,\sigma) \doteq \frac{1}{2}{+}\ln\big(\sqrt{2\pi\sigma^2}\big){-}\mu\:erf\bigg(\frac{\mu}{\sigma\sqrt{2}}\bigg)
  {-}\frac{2\sigma}{\sqrt{2\pi}}e^{-\frac{\mu^2}{2\sigma^2}} \nonumber 
\end{equation}
By the fact that $\frac{d erf(t)}{dt}=\frac{2}{\sqrt{\pi}}e^{-t^2}$ and using standard
calculus of derivatives, we can get that:
\begin{IEEEeqnarray}{ll}
  \frac{\partial H_{B}(\mu,\sigma)}{\partial\mu} &= -erf\bigg(\frac{\mu}{\sigma\sqrt{2}}\bigg) \nonumber\\
  \frac{\partial H_{B}(\mu,\sigma)}{\partial\sigma} &= \frac{1}{\sigma}-\frac{2}{\sqrt{2\pi}}e^{-\frac{\mu^2}{2\sigma^2}} \nonumber
\end{IEEEeqnarray}
where the first line is zero if and only if $\mu=0$, thus proving the first condition in Theorem~\ref{th:stationarity},
while, by setting the second line to zero, we obtain the following equation:
\begin{equation}
  \label{eq:equation}
  \frac{e^{\frac{\mu^2}{2\sigma^2}}}{\sigma}=\frac{2}{\sqrt{2\pi}}
\end{equation}
By the change of variable $t\doteq\frac{\mu}{\sigma\sqrt{2}}$, (\ref{eq:equation}) simplifies
in the following equation:
\begin{equation}
  \label{eq:equation2}
  te^{t^2} = \frac{\mu}{\sqrt{\pi}}
\end{equation}
whose solution is given by:
\begin{equation}
  \label{eq:solution}
  t=\frac{\mu}{\sqrt{\pi}}e^{-\frac{W\big(\frac{2\mu^2}{\pi}\big)}{2}}
\end{equation}
where $W(\cdot)$ is the principal branch of the Lambert W function~\cite{corless1996lambertw}.
We can check whether (\ref{eq:solution}) is the solution of~(\ref{eq:equation2}), by substituting
(\ref{eq:solution}) in~(\ref{eq:equation2}) and see that the equality~(\ref{eq:equation2}) holds,
namely:
\begin{IEEEeqnarray}{ll}
  \frac{\mu}{\sqrt{\pi}}e^{-\frac{W\big(\frac{2\mu^2}{\pi}\big)}{2}}
  e^{\frac{\mu^2}{\pi}e^{-W\big(\frac{2\mu^2}{\pi}\big)}} &= \frac{\mu}{\sqrt{\pi}} \nonumber\\
  e^{-\frac{W\big(\frac{2\mu^2}{\pi}\big)}{2}}
  e^{\frac{\mu^2}{\pi}e^{-W\big(\frac{2\mu^2}{\pi}\big)}} &= 1 \nonumber \\
  e^{-\frac{W\big(\frac{2\mu^2}{\pi}\big)}{2}}
  e^{\frac{2\mu^2}{\pi}\frac{e^{-W\big(\frac{2\mu^2}{\pi}\big)}}{2}} &= 1 \nonumber \\
  e^{-\frac{W\big(\frac{2\mu^2}{\pi}\big)}{2}}
  e^{\frac{W\big(\frac{2\mu^2}{\pi}\big)}{2}} &= 1 \nonumber
\end{IEEEeqnarray}
where the fourth line is obtained by the identity $t=W(t)e^{W(t)}$, or equivalently $te^{-W(t)}=W(t)$.
Therefore, solution~(\ref{eq:solution}) can be explicited in terms of $\sigma$, thus giving us the
second condition in Theorem~\ref{th:stationarity}. 

To prove the optimality of the the obtained solution, we perform a second-order derivative test, namely:
\begin{IEEEeqnarray}{ll}
    \label{eq:secondorder}
    \frac{\partial^2H_B(\mu,\sigma)}{\partial\mu^2} &= -\sqrt{\frac{2}{\pi}}\frac{1}{\sigma}
                                                        e^{-\frac{\mu^2}{2\sigma^2}} \nonumber\\
    \frac{\partial^2H_B(\mu,\sigma)}{\partial\sigma^2} &= - \frac{1}{\sigma^2}-\sqrt{\frac{2}{\pi}}
                                                          \frac{\mu^2}{\sigma^2}e^{-\frac{\mu^2}{2\sigma^2}} \nonumber\\
    \frac{\partial^2H_B(\mu,\sigma)}{\partial\mu\partial\sigma} &= \frac{\partial^2H_B(\mu,\sigma)}{\partial\sigma\partial\mu}
                                    = \sqrt{\frac{2}{\pi}}\frac{\mu}{\sigma^2}e^{-\frac{\mu^2}{2\sigma^2}}
\end{IEEEeqnarray}
Note that the first two equations in~(\ref{eq:secondorder}) are always negative, while the third equation
is zero at $\mu=0$. Therefore the Hessian matrix for the obtained stationary point is negative-definite. \hfill Q.E.D.

{\section{General Case of Different Input Variances}\label{sec:general}}
Suppose that we are given an initial weight vector denoted by $\tilde{\mathbf{w}}\in\mathbb{R}^d$. In general,
$\tilde{\mathbf{w}}$ does not satisy the optimality condition given by Theorem~\ref{th:stationarity}, 
namely $\sum_{i=1}^d\tilde{w}_i^2Var\{x_i\}\neq\pi/2$. We need to find the closest vector 
to $\tilde{\mathbf{w}}$ that satisfies the condition in order to have maximum amount of
information propagation. This can be formulated as the following optimization problem:
\begin{IEEEeqnarray}{ll}
  \label{eq:projection}
  \min_{\mathbf{w}} & \|\mathbf{w}-\tilde{\mathbf{w}}\|_2^2 \nonumber\\
  \text{s.t.} & \mathbf{w}^T\mathbf{D}\mathbf{w}=\pi/2
\end{IEEEeqnarray}
where $\mathbf{D}\doteq diag(Var\{x_1\},\dots,Var\{x_d\})$ and the solution can be obtained 
by using existing iterative non-convex optimization procedures, due to the fact that the 
feasible set in~(\ref{eq:projection}) is not convex.

%
\ifCLASSOPTIONcompsoc
  \section*{Acknowledgments}
\else
  \section*{Acknowledgment}
\fi
We gratefully acknowledge NVIDIA Corporation 
with the donation of a Titan X Pascal machine to support this research. We thank Massimo Zanetti for insightful discussion.

\ifCLASSOPTIONcaptionsoff
  \newpage
\fi

%
\bibliographystyle{IEEEtran}
\bibliography{IEEEabrv,ref}

\end{document}